\documentclass[10pt]{article}
\usepackage{times}
\date{}
\usepackage[
  letterpaper,
  textwidth=5.5in,
  textheight=9in,
  headheight=28pt,
  headsep=12pt
]{geometry}
\usepackage{fancyhdr}
\usepackage[OT1]{fontenc}
\usepackage{graphicx,amsmath,booktabs,array,tabularx}
\usepackage{xcolor,colortbl,textcomp,caption}
\usepackage{capt-of}
\usepackage{placeins,flafter,wrapfig,needspace,natbib,url,hyperref}

\usepackage{amsmath,amsfonts,bm}

\def\eqref#1{equation~\ref{#1}}

\def\1{\bm{1}}

\DeclareMathAlphabet{\mathsfit}{\encodingdefault}{\sfdefault}{m}{sl}
\SetMathAlphabet{\mathsfit}{bold}{\encodingdefault}{\sfdefault}{bx}{n}

\definecolor{HippoEfficiencySummary}{HTML}{F0F4F7}
\newsavebox{\HippoRetryTable}
\newsavebox{\HippoHierarchyBox}
\newsavebox{\HippoMaintenanceBox}

\hypersetup{
    colorlinks=true,
    citecolor=black,  
    linkcolor=blue,  
    urlcolor=blue    
}

\definecolor{HippoModelGray}{HTML}{EBEBEB}
\definecolor{HippoBestBlue}{HTML}{E5EFF9}
\definecolor{HippoSecondYellow}{HTML}{FFF5CF}
\newcommand{\HippoScore}[2]{\mbox{#1\kern1.2pt{\fontsize{8}{9}\selectfont\color{black!55}\textpm\kern.8pt#2}}}
\newcommand{\HippoMean}[1]{#1}
\newcommand{\HippoModelBand}[2]{%
  \multicolumn{#1}{@{}>{\columncolor{HippoModelGray}[0pt][0pt]}c@{}}%
  {\rule[-1.8pt]{0pt}{10.5pt}\textit{#2}}\\}
\newcommand{\HippoModelHead}[1]{\textit{\shortstack[c]{#1}}}
\usepackage{tcolorbox}
\tcbuselibrary{breakable}

\newtcolorbox{hippoexperience}[1]{
  breakable,
  colback=white,
  colframe=black!45,
  colbacktitle=black!5,
  coltitle=black,
  fonttitle=\small\bfseries,
  fontupper=\small,
  boxrule=0.4pt,
  arc=1mm,
  left=8pt,
  right=8pt,
  top=6pt,
  bottom=6pt,
  title={#1},
  title after break={#1\space(continued)},
  lines before break=7,
  before skip=7pt,
  after skip=7pt
}

\newcommand{\HippoBestScore}[2]{\cellcolor{HippoBestBlue}\HippoScore{#1}{#2}}
\newcommand{\HippoSecondScore}[2]{\cellcolor{HippoSecondYellow}\HippoScore{#1}{#2}}
\newcommand{\HippoBestMean}[1]{\cellcolor{HippoBestBlue}\HippoMean{#1}}
\newcommand{\HippoSecondMean}[1]{\cellcolor{HippoSecondYellow}\HippoMean{#1}}

\title{Hippo: Training-Free Agent Self-Evolution via Hippocampal Replay}

\author{
Yutong Song$^{\clubsuit}$ \quad
Jiang Wu$^{\diamondsuit}$ \quad
Yuxuan Ren$^{\heartsuit}$ \quad
Tony Qin$^{\spadesuit}$ \quad
Baoyuan Wang$^{\spadesuit}$%
\thanks{The views expressed in this paper are solely those of the authors
and have not been reviewed or approved by their respective employers.}\\
$^{\clubsuit}$University of California, Irvine \quad
$^{\diamondsuit}$Independent Researcher\\
$^{\heartsuit}$Rice University \quad
$^{\spadesuit}$Zillow Group
}

\makeatletter
\newcommand{\HippoFinishWrap}{%
  \par
  \ifnum\c@WF@wrappedlines>1
    \@tempcnta=\c@WF@wrappedlines
    \advance\@tempcnta by -1
    \dimen@=\baselineskip
    \multiply\dimen@ by \@tempcnta
    \vskip\dimen@
  \fi
  \WFclear
}
\makeatother

\begin{document}

\maketitle

\begin{abstract}
Motivated by recursive self-improvement (RSI), we study how LLM agents can learn through external memory without updating model parameters.
Extracting useful experience from a single attempt is difficult: flawed reasoning can produce success, while poor execution can obscure a sound strategy.
Preserving its value over time is also difficult: our accumulation baseline without consolidation initially improves but eventually performs below a no-memory baseline.
Our key idea is to compare attempts at the same task to identify reusable decisions and refine when they apply.
We introduce Hippo, a memory framework inspired by selective hippocampal encoding and replay that learns from these comparisons when shortcomings are diagnosed.
It compares the resulting experience with existing memory to merge overlaps and resolve conflicts through explicit applicability conditions.
This allows prior guidance to be revised without storing every lesson as a separate entry.
Experiments on SWE-bench Verified, WebArena, and Mind2Web show performance gains in all three settings and substantially fewer retained experiences than the accumulation baseline.
The gains extend across the evaluated base models of varying capabilities, indicating that memory-based adaptation remains useful even for highly capable models.

\end{abstract}
\section{Introduction}
\label{sec:introduction}

Self-evolving agents seek to turn experience from past tasks into better decisions on future tasks~\citep{gao2026survey}.
This objective has motivated research on recursive self-improvement, including agents that revise their own code in response to execution feedback~\citep{zhang2025darwin}.
External memory offers a form of \emph{gradient-free learning} without changes to model parameters~\citep{shinn2023reflexion,zhao2024expel}.
Past experience becomes part of the context for later decisions, so the agent's adaptation depends on what it retains and how it revises that experience.

Prior work has established several ways to represent, update, and reuse experience.
ReasoningBank's MaTTS also compares attempts to form memory, while its reported consolidation implementation appends new items without further pruning or merging~\citep{ouyang2026reasoningbank}.Vanilla-IB serves as a baseline for examining how interaction steps relate to token usage and task performance~\citep{hajimiri2026budget}.We focus on an update policy in which identified shortcomings determine whether to form new experience, what it should contain, and how it should revise prior guidance.

\begin{figure}[!htbp]
    \centering
    \includegraphics[width=\linewidth]{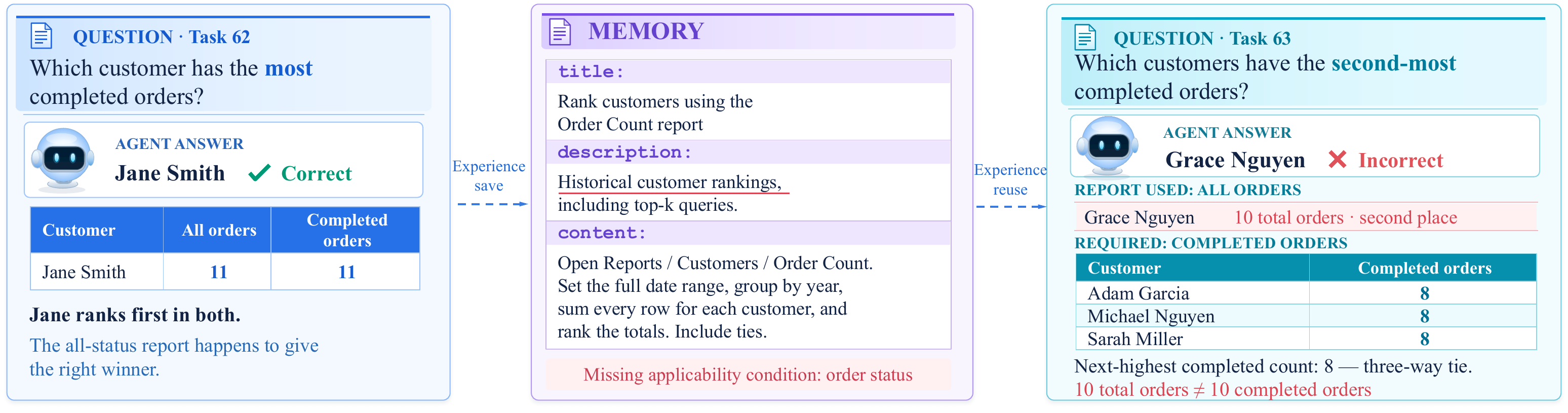}
    \caption{
        Correct outcomes can mask flawed reasoning, leading to erroneous
        or overly general lessons. Once retained in memory, these lessons
        may be repeatedly retrieved and misguide decisions across
        subsequent tasks.
    }
    \label{fig:motivation}
\end{figure}

Continued adaptation, however, does not per se guarantee continued improvement. An exmaple is illustrated in Figure \ref{fig:motivation}.
In our experiments, we examine an accumulation baseline that adds experience from both successful and failed attempts without consolidation.
With this baseline, performance follows an inverted-U pattern as memory grows: it improves at first, then declines and eventually falls below the no-memory baseline.

This observation motivates our central question: \emph{how should an agent decide what experience to retain and how that experience should revise existing memory?}

An execution trace does not directly reveal what the agent should learn from it.
A single failed attempt may reflect a flawed strategy or poor execution of a valid plan.
A correct outcome can also conceal incorrect reasoning.
A reflection may therefore extract a flawed lesson that later retrieval repeatedly presents as a trusted precedent.
Even a sound lesson needs to capture the decision behind the observed actions and the conditions under which that decision applies.
Task-specific action sequences may transfer poorly, while generic advice may offer little practical guidance.
New information may also duplicate or contradict what the agent already retains~\citep{chhikara2025mem0}.

We propose Hippo, a memory framework for self-evolving agents.
The design draws on selective hippocampal encoding of salient episodes and replay of past experience during consolidation~\citep{lisman2005hippocampus,kumaran2016learning}.
These processes motivate a separation between episodic reflections on individual attempts and persistent agent memory that integrates experience across tasks.
Hippo reflects only on attempts judged to fail or succeed for the wrong reasons.
We refer to these cases as \emph{surprise}.

Hippo uses two comparisons to determine how memory should change.
It first compares multiple attempts at the same task, sampled under the same frozen policy and retrieved memory, to distill candidate experience from their decisions and outcomes.
It then compares this experience with related memories to merge overlaps and resolve conflicts through explicit applicability conditions.
These comparisons determine both what can be learned from the current task and how it should revise prior guidance.
A revision can therefore change existing guidance without adding another entry.
The revised memory guides decisions on subsequent tasks.
Experience from those tasks, in turn, provides evidence for further memory updates.

We evaluate Hippo on software issue resolution with SWE-bench Verified~\citep{jimenez2024swebench,openai2024swebenchverified}, interactive web task completion in WebArena~\citep{zhou2024webarena}, and next-action prediction from recorded webpage states in Mind2Web~\citep{deng2023mind2web}.
Hippo improves task performance in all three settings and retains substantially fewer experiences than the accumulation baseline.
Experiments show gains across base models with different levels of capability.
These gains extend to highly capable models such as GPT-5.6, which indicates that memory-based adaptation remains useful even for agents with strong base capabilities. Our main contributions are:


\begin{itemize}
    \item We quantify how repeated attempts improve the chance
    of task success even without memory updates. This insight
    motivates direct retries as a baseline for future
    evaluations of self-evolving agents under comparable
    budgets, feedback, and success criteria.

    \item We propose Hippo, a framework for memory-based
    self-evolution that couples selective experience induction
    from multiple attempts at the same task with maintenance
    of existing memory. Experiments across task settings and
    base models demonstrate its effectiveness.

    \item We identify a failure mode of memory-based
    self-evolution: continued experience accumulation without
    consolidation can erode initial gains and eventually
    reduce performance below the no-memory baseline.
\end{itemize}
\section{Related Works}
\label{sec:related_work}

Generative Agents ~\citep{park2023generative} use reflection to derive higher-level memories from stored observations.
Reflexion records reflections on task feedback for subsequent trials~\citep{shinn2023reflexion}, while Agent Workflow Memory abstracts recurring routines into reusable workflows~\citep{wang2025awm}.
However, \citet{zhao2026faithful} find that task success does not consistently decline when condensed experience is replaced with meaningless symbols.
This result also holds in their ReasoningBank evaluation, where condensed experience is \textit{de facto} the sole source of historical task information available to the agent.

ReasoningBank's parallel MaTTS compares multiple trajectories for the same query to distill reasoning strategies~\citep{ouyang2026reasoningbank}.
Hippo ties experience induction to task-level prediction gap and to distill and revise semantic knowledge.
MemGPT manages information across in-context and external storage~\citep{packer2023memgpt}.
ExpeL revises or removes stored insights~\citep{zhao2024expel}.
Mem0 extracts and updates salient conversational information~\citep{chhikara2025mem0}, while A-MEM updates the descriptions and attributes of related memories~\citep{xu2025amem}.
ReasoningBank's reported consolidation implementation directly appends new entries without further pruning or merging~\citep{ouyang2026reasoningbank}.
MemRL uses runtime reinforcement learning to refine retrieval utilities while its LLM remains frozen~\citep{zhang2026memrl}.
Hippo compares new experience with related memories to merge overlaps and resolve conflicts through explicit applicability conditions.
Memory-R1 and MEM1 use reinforcement learning to train memory policies~\citep{yan2025memoryr1,zhou2025mem1}.
Hippo instead updates external memory while model parameters fixed.
\section{Hippo}
\begin{figure}[t]
    \centering
    \includegraphics[width=\linewidth]{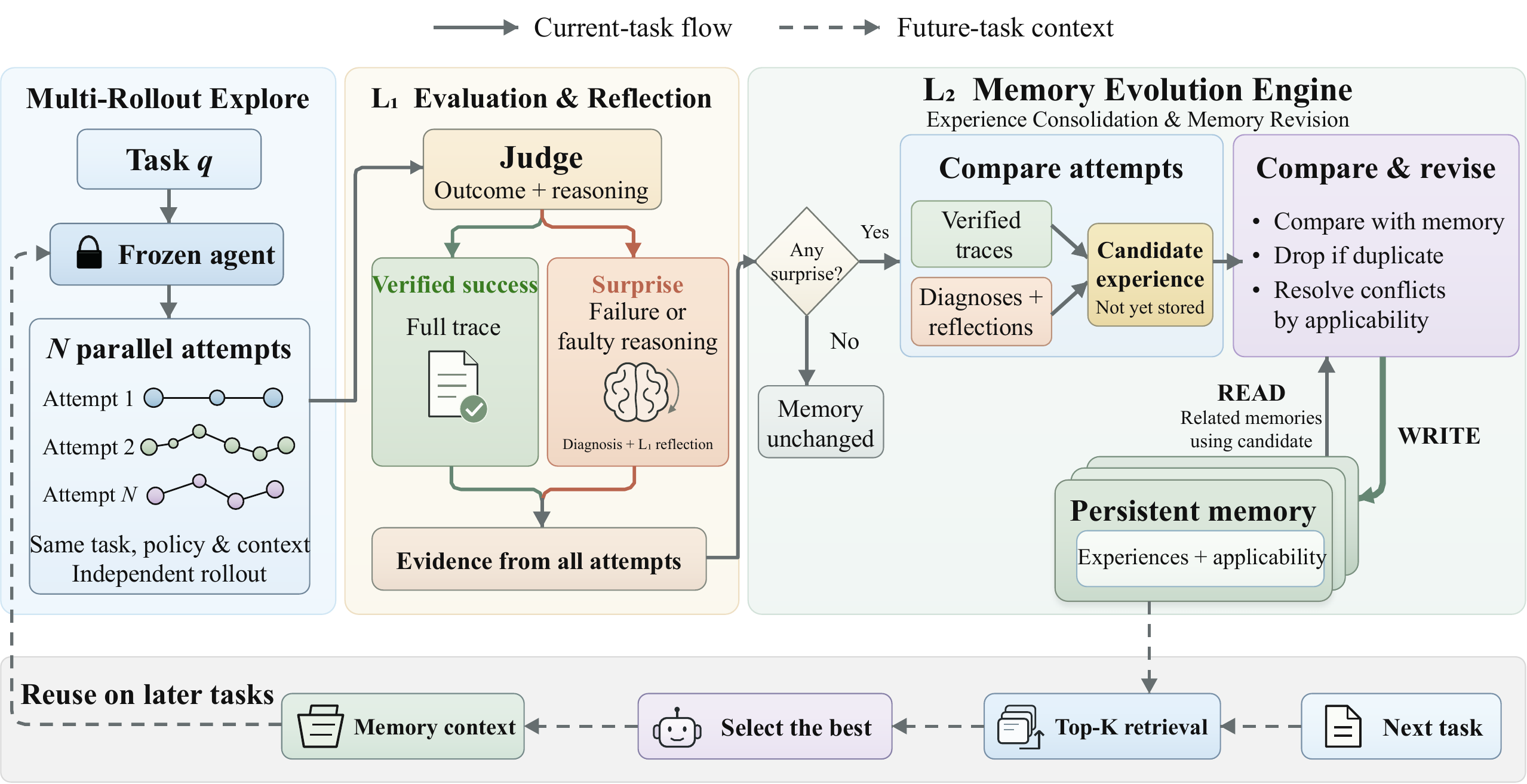}
    \caption{
Overview of Hippo.
    Hippo combines surprise-driven episodic reflection with memory
    consolidation inspired by hippocampal replay, while keeping
    the agent policy fixed.
}
    \label{fig:hippo}
\end{figure}

Architecturally, Hippo updates its memory from multiple attempts at the same task through four components. (i) The multi-rollout exploration module runs the agent \(N\) times with the same frozen policy and retrieved memory. (ii) The surprise-driven episodic buffer records reflections from attempts that fail or succeed for the wrong reasons, acting as a hippocampal episodic buffer. (iii) The memory evolution engine draws on hippocampal replay to transform these reflections into the final agent memory. It compares the rollouts to distill reusable experience, then checks this experience against existing memory to merge overlaps and resolve conflicts. (iv) The retrieval and injection module selects relevant memories and provides them to the agent for later tasks. Figure \ref{fig:hippo} shows the overall architecture.
\subsection{Surprise-Driven Episodic Reflection}
\label{sec:episodic_buffer}

A core challenge in self-evolving memory systems is to decide what content deserves retention.
Models of the hippocampal--VTA loop link novelty signals to enhanced long-term encoding~\citep{lisman2005hippocampus}, while human imaging studies show hippocampal responses to violations of learned event sequences~\citep{kumaran2006mismatch}.
In self-evolving systems, simply accumulating routine experiences often steers the agent off-course and degrades performance over time ((Section~\ref{sec:long-horizon-performance})).
We use a judge $\mathcal{J}$ to evaluate each rollout and determine whether to extract an episodic reflection:
\begin{equation}
\label{eq:episodic_buffer}
    L_1 = \begin{cases}
        \mathcal{G}(q, \tau, \delta), & \text{if } \mathcal{J}(q, \tau, r) = \text{surprise}, \\
        \emptyset, & \text{otherwise}.
    \end{cases}
\end{equation}
Here, $q$ is the task, $\tau$ is the execution trace. Here, $r$ denotes feedback when available
and is empty otherwise. On SWE-bench Verified, the judge estimates
success from the execution trace. $\delta$ is the diagnosis returned by $\mathcal{J}$, and $\mathcal{G}$ is the reflection extractor.
The judge returns surprise unless the task succeeds and the agent's reasoning is correct.
Thus, rollouts that are right for the wrong reasons still count as surprises and yield episodic reflections $L_1$.
\subsection{Multi-Rollout Exploration Scaffold}
\label{sec:exploration_scaffold}

A single failed trajectory may not reveal whether the approach was flawed or a valid plan was poorly executed.
Models of causal induction emphasize comparisons across relevant cases~\citep{cheng1990probabilistic}.
 Multiple attempts under the same conditions provide a broader evidence base and reduce the risk of overfitting a lesson to one trajectory.
We therefore apply the episodic buffer of Section~\ref{sec:episodic_buffer} to $N$ rollouts of the same task, sampled in parallel from the frozen agent policy $\pi$ with temperature sampling and conditioned on the same retrieved memory $\mathcal{M}$:
\begin{equation}
    \mathcal{E}(q) = \big\{ (\tau_i, r_i, \delta_i, L_1^{(i)}) \big\}_{i=1}^{N}, \qquad \tau_i \sim \pi(\cdot \mid q, \mathcal{M}).
\end{equation}
Each rollout is judged and reflected on independently by Eq.~\eqref{eq:episodic_buffer}, and executes concurrently in its own replica of the space so that the $N$ outcomes remain independent measurements. 
Together, these rollouts provide broader experience and isolate patterns that generalize across trajectories. 
The resulting experience set $\mathcal{E}(q)$ serves as the input to the memory evolution engine, which produces the final agent memory $L_2$ (Section~\ref{sec:memory_evolution}).

\subsection{Memory Evolution Engine}
\label{sec:memory_evolution}

A central challenge for self-evolving agent memory is to remain useful over long task streams.
Our experiments show that evaluated basedline performance follows an inverted-U curve as memory grows: it improves at first, then falls below the no-memory baseline (Section~\ref{sec:long-horizon-performance}).
We further observe three problems: (i) (i) a single reflection can extract a flawed lesson from limited
evidence, and repeated retrieval may carry that error into later tasks;
(ii) reflections may retain surface actions rather than the decisions behind them, and (iii) duplicate or conflicting experiences make retrieval unreliable.
Thus, For a self-evolving agent that continuously tackle new tasks, it requires an active mechanism to revise, merge, and prune past experiences.

Hippocampal replay provides biological inspiration for this process.
\citet{wilson1994reactivation} show that hippocampal activity patterns associated with waking experience can reactivate during subsequent sleep.
In complementary learning systems theory, reactivation supports the gradual integration of new experience with existing neocortical knowledge~\citep{mcclelland1995cls,kumaran2016learning}.
This account inspires Hippo's two-stage update.
It first contrasts the $N$ rollouts in $\mathcal{E}(q)$ to distill a candidate experience.
It then compares the candidate against existing memory to merge overlapping experiences and resolve conflicts.
Together, these stages transform episodic reflections $L_1$ into the final agent memory $L_2$.

\paragraph{Contrastive Experience Induction.}
We compare attempts at the same task to derive a candidate
experience $m_t$: a decision together with its conditions of use.
The induction operator $\mathcal{D}$ performs this comparison
with an LLM after selecting the evidence supplied by each rollout.

A verified success has a successful outcome and reasoning
accepted by the judge.
Let $n_c$ count verified successes among the $N$ attempts
at $q_t$, and set $\rho_t=n_c/N$.
From $\mathcal{E}(q_t)$, $\mathcal{D}$ selects full traces
for verified successes and judge diagnoses with $L_1$ reflections
for all other attempts, including successes for the wrong reasons.
For $\rho_t<1$, the process is
\begin{equation}
\label{eq:contrastive_induction}
    \bigl(q_t,\mathcal{E}(q_t),\rho_t\bigr)
    \xrightarrow{\;\mathcal{D}\;}
    \underbrace{m_t}_{\text{candidate experience}}.
\end{equation}
The LLM is not given the ground-truth target directly.
When $\rho_t=1$, induction is skipped and memory remains unchanged.

For batches that trigger induction, $\mathcal{D}$ follows
four cases.
(i) With a verified-success majority, it turns the minority's
errors into cautions.
(ii) With a nonempty verified-success minority, it identifies
decisions that distinguish successful attempts.
(iii) With an even split, it extracts both a successful
strategy and a failure trap.
(iv) With no verified successes, it summarizes failure patterns
documented in the diagnoses and reflections.
In the last case, it is instructed to restrict guidance
to actions supported by those inputs.
The resulting candidate is passed to memory consolidation.

\paragraph{Memory Consolidation.}
Let $L_{2,t}$ be the persistent $L_2$ bank, represented as a set
of identified entries, and $\mathcal{R}_t\subseteq L_{2,t}$ the
entries retrieved by similarity to a nonempty candidate $m_t$.
The language-model comparator $\mathcal{C}$ checks for redundancy
and conflicting guidance, then specifies additions $\Delta_t^{+}$
and deletions $\Delta_t^{-}\subseteq\mathcal{R}_t$:
\begin{equation}
\label{eq:memory_consolidation}
\begin{aligned}
    (\Delta_t^{+},\Delta_t^{-})
        &=\mathcal{C}(m_t,\mathcal{R}_t),\\
    L_{2,t+1}
        &=\underbrace{
            (L_{2,t}\setminus\Delta_t^{-})
          }_{\text{unchanged entries}}
          \cup
          \underbrace{
            \Delta_t^{+}
          }_{\text{new entries}}.
\end{aligned}
\end{equation}
Duplicate candidates are discarded.
Overlaps and conflicts are addressed by making applicability
conditions explicit in the new entries.
When new guidance replaces an old entry, the engine adds the
replacement and deletes the old entry.
Entries outside the deletion set retain their original text.
Section~\ref{sec:ablation} compares this update policy
with append-only updates and in-place rewriting.


\subsection{Memory Retrieval and Injection}
\label{sec:retrieval_injection}

Semantic similarity helps identify related experiences, but does not determine whether their guidance applies to the current task.
Hippo therefore uses a two-stage procedure that combines embedding retrieval with agent-based selection.
For each task $q$, we first retrieve the top-$K$ candidate experiences from $L_2$ by embedding similarity.
The agent then evaluates these candidates against the task requirements and selects the applicable experiences as $\mathcal{M}$.
We inject $\mathcal{M}$ into the agent's context before execution, and all $N$ rollouts of $q$ share the same selected memory.
At each step, the agent selects an action under the frozen policy $\pi$, conditioned on the current task state and $\mathcal{M}$.

We compare this procedure with BM25~\citep{robertson2009bm25}, BGE-based top-$K$ retrieval~\citep{xiao2024cpack}, hybrid sparse--dense retrieval via reciprocal rank fusion~\citep{cormack2009reciprocal}, and direct LLM-based memory selection inspired by RankGPT~\citep{sun2023rankgpt}.
On recorded WebArena queries with a matching experience
available, Hippo achieves a selection rate of 75.2\%.
Appendix~\ref{app:hippo_retrieval} provides the evaluation
protocol and detailed comparisons.

\section{Experiments and Results Analysis} 
\label{sec:experiments}

\label{sec:experimental_setup}

\paragraph{Benchmarks and models.}
We evaluate Hippo on SWE-bench Verified~\citep{jimenez2024swebench,openai2024swebenchverified} for coding task, WebArena~\citep{zhou2024webarena} for interactive web tasks, and Mind2Web~\citep{deng2023mind2web} for offline next-action prediction.
We use four base models with fixed parameters: GPT-5.6 Sol, Claude Haiku 4.5, Gemini 3 Flash Preview, and Qwen3.5-27B.
The result tables list the models used for each benchmark.

\paragraph{Baselines.}
We compare Hippo with No Memory,
Vanilla-IB~\citep{hajimiri2026budget},
AWM~\citep{wang2025awm}, and
ReasoningBank~\citep{ouyang2026reasoningbank}.
No Memory and Vanilla-IB operate without memory.
We use the official implementations of AWM and
ReasoningBank and enable MaTTS for ReasoningBank.
On WebArena, both methods receive official outcomes
and reference answers during memory construction.
All methods and ablated variants use at most
30 interaction steps per rollout, except Vanilla-IB,
which uses 50.
Appendix~\ref{app:surprise_protocol} describes feedback
access, and Appendix~\ref{sec:appendix_efficiency}
reports interaction lengths.

\paragraph{Evaluation.}
We report resolution rate on all 500 SWE-bench Verified tasks, task success rate on WebArena, and step success rate on Mind2Web.
For the main results, we average scores over eight attempts per task, then across tasks. We report the mean $\pm$ sample standard deviation across
three complete runs with different random seeds.
Further details are provided in Appendix~\ref{app:experimental_details}.

\subsection{Performance among different benchmarks}

\begin{table}[!htbp]
\centering
\small
\caption{Performance comparison on WebArena. Blue and yellow highlight the best and second-best results, respectively, in each column within each model group.}
\vspace{-1em}
\begingroup
\fontfamily{ptm}\selectfont
\fontsize{10}{11}\selectfont
\setlength{\tabcolsep}{4pt}
\renewcommand{\arraystretch}{0.86}
\setlength{\heavyrulewidth}{.8pt}
\setlength{\lightrulewidth}{.4pt}
\setlength{\aboverulesep}{1pt}
\setlength{\belowrulesep}{1pt}
\begin{tabular*}{\linewidth}{@{\hspace{4pt}}l@{\extracolsep{\fill}}cccccc@{\hspace{4pt}}}
\toprule
\textbf{Method} & \textbf{Shopping} & \textbf{Admin} & \textbf{GitLab} & \textbf{Reddit} & \textbf{Map} & \textbf{Overall} \\
 & (187) & (182) & (180) & (106) & (109) & (764) \\
\midrule
\HippoModelBand{7}{GPT-5.6 Sol}
No Memory & \HippoScore{53.4}{1.1} & \HippoScore{78.9}{0.6} & \HippoScore{67.9}{0.8} & \HippoScore{56.8}{1.2} & \HippoScore{59.3}{1.2} & \HippoMean{64.2} \\
Vanilla-IB & \HippoScore{54.0}{0.6} & \HippoSecondScore{80.5}{1.2} & \HippoSecondScore{68.5}{1.0} & \HippoScore{57.6}{0.8} & \HippoSecondScore{60.1}{0.8} & \HippoSecondMean{65.1} \\
AWM & \HippoScore{52.7}{3.4} & \HippoScore{78.7}{3.4} & \HippoScore{67.7}{3.0} & \HippoSecondScore{57.8}{2.1} & \HippoScore{58.8}{2.3} & \HippoMean{64.0} \\
ReasoningBank & \HippoSecondScore{54.3}{1.3} & \HippoScore{79.2}{2.5} & \HippoScore{67.7}{3.2} & \HippoScore{56.3}{2.8} & \HippoScore{59.1}{3.2} & \HippoMean{64.4} \\
\addlinespace[0pt]
Hippo & \HippoBestScore{64.2}{0.6} & \HippoBestScore{87.2}{1.3} & \HippoBestScore{73.3}{3.0} & \HippoBestScore{62.5}{0.8} & \HippoBestScore{73.4}{1.0} & \HippoBestMean{72.9} \\
\midrule
\HippoModelBand{7}{Claude Haiku 4.5}
No Memory & \HippoScore{39.2}{1.2} & \HippoScore{31.4}{1.0} & \HippoScore{33.1}{0.8} & \HippoScore{34.8}{0.6} & \HippoScore{28.7}{0.6} & \HippoMean{33.8} \\
Vanilla-IB & \HippoSecondScore{40.8}{1.2} & \HippoSecondScore{37.0}{0.8} & \HippoSecondScore{34.6}{1.2} & \HippoScore{36.4}{1.0} & \HippoSecondScore{30.8}{0.7} & \HippoSecondMean{36.4} \\
AWM & \HippoScore{38.5}{3.8} & \HippoScore{29.9}{3.8} & \HippoScore{32.9}{3.4} & \HippoSecondScore{36.7}{4.4} & \HippoScore{27.4}{2.2} & \HippoMean{33.3} \\
ReasoningBank & \HippoScore{39.8}{2.2} & \HippoScore{33.2}{2.2} & \HippoScore{33.6}{2.9} & \HippoScore{35.3}{2.8} & \HippoScore{29.0}{2.9} & \HippoMean{34.6} \\
\addlinespace[0pt]
Hippo & \HippoBestScore{50.1}{1.5} & \HippoBestScore{39.8}{1.0} & \HippoBestScore{38.6}{1.0} & \HippoBestScore{40.6}{1.2} & \HippoBestScore{42.3}{0.8} & \HippoBestMean{42.5} \\
\midrule
\HippoModelBand{7}{Gemini 3 Flash Preview}
No Memory
& \HippoScore{46.2}{0.4}
& \HippoScore{48.3}{1.0}
& \HippoScore{26.8}{1.2}
& \HippoScore{44.3}{0.7}
& \HippoScore{26.3}{1.0}
& \HippoMean{39.0} \\

Vanilla-IB
& \HippoScore{46.2}{1.2}
& \HippoSecondScore{54.3}{1.7}
& \HippoSecondScore{29.1}{1.3}
& \HippoScore{47.4}{0.8}
& \HippoScore{29.5}{0.5}
& \HippoSecondMean{41.9} \\

AWM
& \HippoScore{42.0}{1.2}
& \HippoScore{46.7}{0.5}
& \HippoScore{24.7}{1.2}
& \HippoScore{46.5}{1.7}
& \HippoScore{24.1}{1.8}
& \HippoMean{37.1} \\

ReasoningBank
& \HippoSecondScore{46.3}{0.3}
& \HippoScore{49.3}{0.6}
& \HippoScore{23.9}{0.4}
& \HippoSecondScore{48.3}{0.6}
& \HippoSecondScore{30.9}{0.5}
& \HippoMean{39.8} \\

\addlinespace[0pt]
Hippo
& \HippoBestScore{53.8}{0.6}
& \HippoBestScore{56.9}{0.8}
& \HippoBestScore{32.5}{0.5}
& \HippoBestScore{50.8}{0.5}
& \HippoBestScore{34.8}{0.8}
& \HippoBestMean{46.4} \\
\midrule

\HippoModelBand{7}{Qwen3.5-27B}
No Memory & \HippoScore{42.0}{1.0} & \HippoScore{41.7}{0.5} & \HippoScore{24.3}{0.4} & \HippoScore{40.1}{0.8} & \HippoScore{30.1}{1.2} & \HippoMean{35.8} \\
Vanilla-IB & \HippoSecondScore{43.5}{0.6} & \HippoSecondScore{47.4}{1.3} & \HippoSecondScore{25.9}{0.7} & \HippoScore{41.8}{0.6} & \HippoScore{31.8}{1.1} & \HippoSecondMean{38.4} \\
AWM & \HippoScore{41.1}{3.6} & \HippoScore{40.2}{3.7} & \HippoScore{24.0}{3.6} & \HippoScore{42.0}{4.2} & \HippoScore{30.0}{2.6} & \HippoMean{35.4} \\
ReasoningBank & \HippoSecondScore{43.5}{1.0} & \HippoScore{43.3}{2.1} & \HippoScore{24.9}{2.1} & \HippoSecondScore{42.4}{2.6} & \HippoSecondScore{31.9}{0.9} & \HippoMean{37.3} \\
\addlinespace[0pt]
Hippo & \HippoBestScore{51.7}{1.2} & \HippoBestScore{49.2}{1.0} & \HippoBestScore{29.2}{2.3} & \HippoBestScore{45.2}{1.0} & \HippoBestScore{42.6}{0.7} & \HippoBestMean{43.6} \\
\bottomrule
\end{tabular*}
\endgroup
\label{webarena}
\end{table}

\begin{wraptable}{r}{0.65\textwidth}
\centering
\caption{Performance comparison on SWE-bench.
Blue and yellow highlight the best and second-best results,
respectively, in each model column.}
\vspace{-1em}
\label{swe-bench}
\begingroup
\fontfamily{ptm}\selectfont
\fontsize{10}{11}\selectfont
\setlength{\tabcolsep}{2pt}
\renewcommand{\arraystretch}{0.72}
\setlength{\heavyrulewidth}{.8pt}
\setlength{\lightrulewidth}{.4pt}
\setlength{\aboverulesep}{0.4pt}
\setlength{\belowrulesep}{0.6pt}

\begin{tabular}{@{}lcccc@{}}
\toprule
\rowcolor{HippoModelGray}
\textbf{Method}
& \HippoModelHead{GPT-5.6\\[-1pt]Sol}
& \HippoModelHead{Claude\\[-1pt]Haiku 4.5}
& \HippoModelHead{Gemini 3 Flash\\[-1pt]Preview}
& \HippoModelHead{Qwen3.5\\[-1pt]27B} \\
\midrule
No Memory
& \HippoScore{81.5}{0.4}
& \HippoScore{62.6}{0.2}
& \HippoScore{74.2}{0.4}
& \HippoScore{68.8}{0.6} \\

Vanilla-IB
& \HippoSecondScore{82.6}{0.6}
& \HippoSecondScore{63.2}{0.2}
& \HippoSecondScore{75.2}{0.8}
& \HippoScore{69.2}{0.5} \\

AWM
& \HippoScore{79.9}{1.5}
& \HippoScore{61.0}{2.5}
& \HippoScore{73.7}{1.9}
& \HippoScore{67.4}{2.4} \\

ReasoningBank
& \HippoScore{82.5}{1.9}
& \HippoScore{62.8}{1.5}
& \HippoScore{75.0}{2.1}
& \HippoSecondScore{70.1}{2.4} \\
\midrule
Hippo
& \HippoBestScore{88.5}{1.2}
& \HippoBestScore{70.7}{1.4}
& \HippoBestScore{81.6}{1.4}
& \HippoBestScore{76.5}{1.8} \\
\bottomrule
\end{tabular}
\endgroup
\label{swe-bench}
\end{wraptable}

\begin{wrapfigure}{r}{0.44\textwidth}
    \centering
    \includegraphics[width=\linewidth]
        {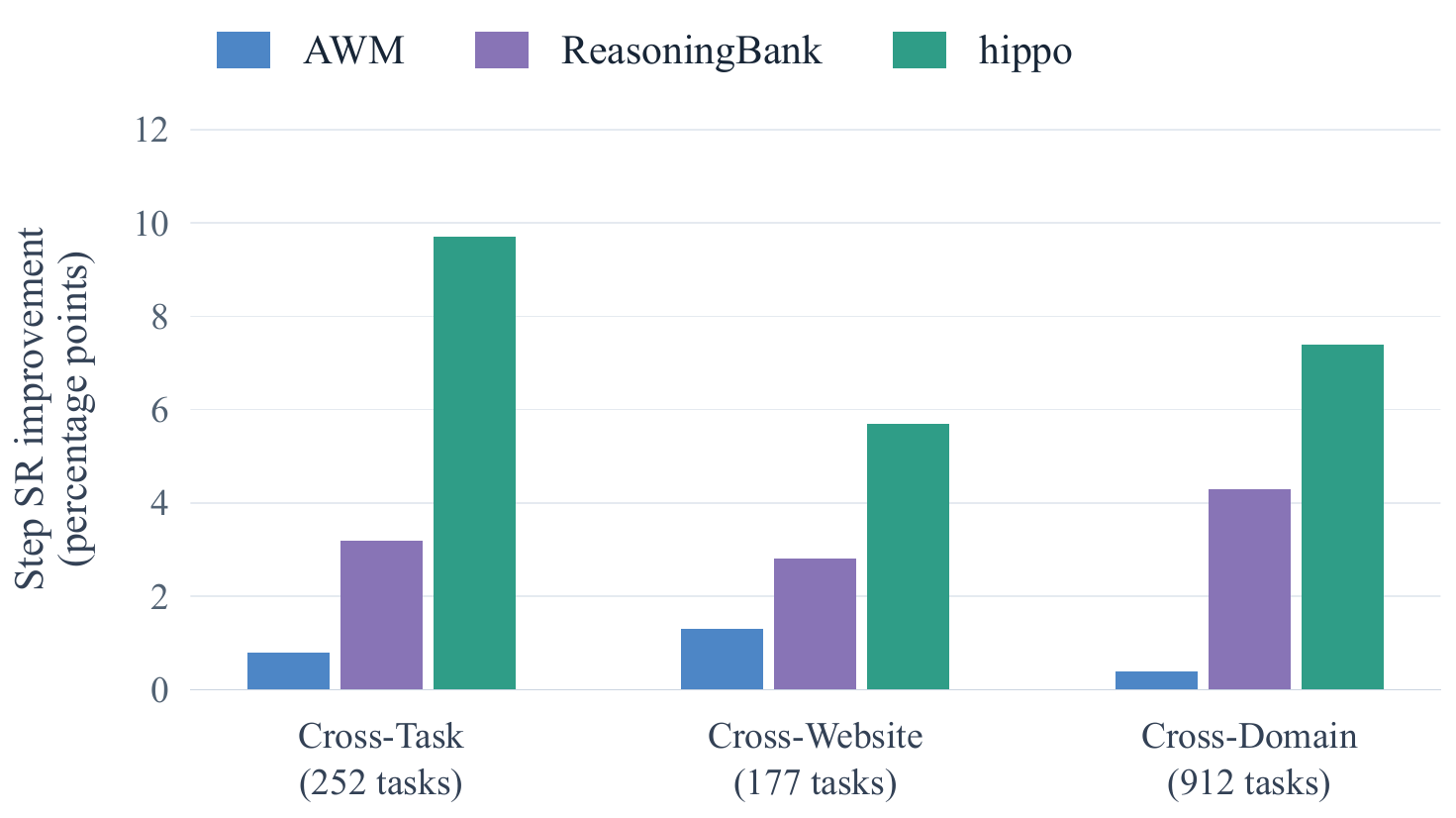}
    \captionof{figure}{Improvements in step success rate over the No Memory baseline for different methods across the three Mind2Web evaluation splits.}
    \label{fig:mind2web}
\end{wrapfigure}

Hippo achieves the best overall performance across all evaluated backbones on WebArena in Table \ref{webarena} and SWE-bench in Table \ref{swe-bench}, and leads across all three generalization splits on Mind2Web in Figure \ref{fig:mind2web}. The benefits of AWM and ReasoningBank vary across models and task environments, indicating that adding memory does not necessarily improve performance. The more consistent gains observed with Hippo align with its design goal of improving experience reuse through selective induction and applicability control.

The task-level results on WebArena reveal substantial variation in the benefits of memory augmentation. Hippo generally achieves larger gains on Shopping and Map, while improvements on GitLab and Reddit are more modest. Despite these differences, Hippo outperforms No Memory on every website across all evaluated backbones and surpasses the other memory methods in most settings.

On Mind2Web, Hippo leads across the Cross-Task, Cross-Website, and Cross-Domain splits with Gemini 3 Flash Preview. We omit further memory-augmented comparisons with GPT-5.6 Sol because its No Memory performance already exceeds 90\% under our evaluation setting.

Notably, even with GPT-5.6 Sol, Hippo improves the overall WebArena score by $8.7\%$ and outperforms No Memory on all five websites. It also improves this backbone's performance on SWE-bench, alongside consistent gains across the other evaluated models. These results indicate that strong model capabilities still leave room for improvement through experience reuse, providing empirical support that Hippo can learn useful guidance from interaction histories and improve agent performance without updating model parameters.

\Needspace{190pt}


\subsection{Performance during Continued Learning}
\label{sec:long-horizon-performance}

Long-term self-evolution requires experience that remains useful
across many tasks and repeated memory updates, growing memory bank alone does not demonstrate continued learning..
Prior work shows that later adaptation can weaken earlier
capabilities despite fixed model parameters~\citep{yu2026self}.
We investigate this behavior using GPT-5.6 Sol on a collection
of 450 tasks from SWE-bench Verified and
SWE-bench Pro~\citep{deng2025swebenchpro},
with all methods following the same task order.
At each checkpoint, we evaluate performance against the
no-memory baseline on the first 50 tasks and the same fixed
set of 50 unseen tasks, assessing retention and transfer,
respectively.
Evaluation does not modify the memory used for subsequent learning.

\begingroup
\setlength{\intextsep}{6pt}
\begin{wrapfigure}{r}{0.70\textwidth}
    \centering
    \captionsetup{
        font=footnotesize,
        skip=4pt,
        justification=raggedright,
        singlelinecheck=false
    }
    \includegraphics[
        width=\linewidth,
        trim={0 7 0 8},
        clip
    ]{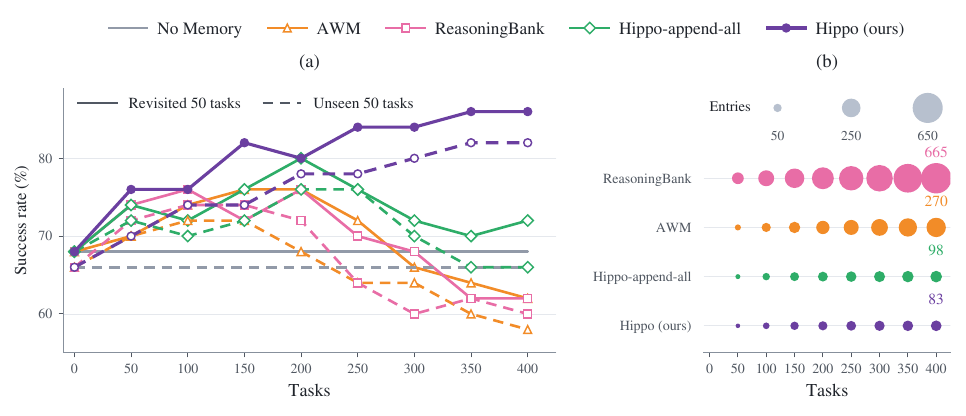}

    \caption{Long-horizon performance and memory size
        with GPT-5.6 Sol.
        (a) Success rates on the first 50 tasks and a fixed
        set of 50 unseen tasks, evaluated at each checkpoint.
        (b) Retained memory entries, indicated by circle size.
    }
    \label{fig:continual-learning}
\end{wrapfigure}


AWM and ReasoningBank initially benefit from memory, but
eventually underperform their corresponding no-memory
baselines on both task sets
(Figure~\ref{fig:continual-learning}(a)).
This decline occurs despite continued growth in their
memory banks (Figure~\ref{fig:continual-learning}(b)).
In this sequence, performance crosses below the baseline
earlier on unseen tasks than on the first 50 tasks.

Two possible contributing factors: reduced access to useful
experience and repeated use of misleading experience.
First, newly appended entries may compete with earlier
entries for retrieval and context space, leaving useful
experience stored but unavailable to the agent during execution.
Second, self-evaluation may overrate flawed trajectories,
and experience induction may produce incorrect rules or
omit their conditions of use.
Repeated retrieval of these entries could then propagate
misleading guidance to later tasks.

Hippo sustains gains over the no-memory baseline on both
early and unseen tasks and achieves the highest final
performance among the evaluated methods.
At the 400-task checkpoint, success rates reach
86\% and 82\%, respectively, exceeding the strongest
memory baseline on each task set by 24 and 22 percentage points.
At the same checkpoint, Hippo retains 83 entries, compared
with 270 for AWM and 665 for ReasoningBank
(Figure~\ref{fig:continual-learning}(b)).
These results show sustained gains with fewer stored
entries in this task sequence. Also, at the 400-task checkpoint, full Hippo outperforms its
append-only variant by 14.0 percentage points on the revisited
tasks and 16 percentage points on the unseen tasks
(Table~\ref{tab:long_append_only}).

\HippoFinishWrap
\endgroup
\FloatBarrier
\subsection{Exploration Before Memory Updates}
\label{sec:exploration-before-memory}


\textbf{More steps.}\enspace
\citet{hajimiri2026budget} show that more interaction steps can improve task success without a memory system.
We observe the same pattern: with GPT-5.6 Sol, Vanilla-IB gives the agent a larger step budget and improves success rates over No Memory by an average of 0.9 \% across four WebArena domains. 

\noindent\textbf{More tries.}\enspace
Our experiments reveal a simple insight: \emph{more attempts alone can improve task success}.
A method may spend extra computation updating its harness or memory from one trajectory, yet score can lower than the original agent just given more attempts.
This motivates an evaluation criterion: a self-evolving method must outperform direct retries under the feedback conditions, and success criterion.
Otherwise, the evidence is insufficient to support its claimed effectiveness.

Averaged equally across four WebArena websites,
No Memory achieves $64.3\%$ mean per-attempt success
and $73.6\%$ success at least once within eight attempts,
a difference of $9.3$ percentage points.
Appendix~\ref{sec:retry_metrics} provides the metric
definitions and normalized gains.

\HippoFinishWrap
\Needspace{165pt}
\begin{wraptable}{r}{0.4\textwidth}
    \centering
    \captionsetup{
        position=top,
        font=footnotesize,
        skip=4pt,
        justification=raggedright,
        singlelinecheck=false
    }
    \setlength{\tabcolsep}{1.5pt}
    \renewcommand{\arraystretch}{1.07}
    \setlength{\aboverulesep}{2pt}
    \setlength{\belowrulesep}{2pt}

    \captionof{table}{%
        \textbf{More tries.}
        For success rates $s_1,s_n\in[0,1]$,
        success gain is $\Delta_n=s_n-s_1$.
        Recovery rate is the fraction of mean per-attempt
        failures solved by later attempts:
        $R_n=\frac{s_n-s_1}{1-s_1}$ ($n=8$).
    }
    \label{tab:retry-benefits}

    \fontsize{8.5}{10}\selectfont
    \resizebox{\linewidth}{!}{%
    \begin{tabular}{@{}lrrrr@{}}
        \toprule
        & \multicolumn{1}{c}{Single}
        & \multicolumn{1}{c}{Any success}
        & \multicolumn{1}{c}{Success}
        & \multicolumn{1}{c@{}}{Recovery} \\
        Domain
        & \multicolumn{1}{c}{attempt}
        & \multicolumn{1}{c}{attempts}
        & \multicolumn{1}{c}{gain}
        & \multicolumn{1}{c@{}}{rate} \\
        & \multicolumn{1}{c}{(\%)}
        & \multicolumn{1}{c}{(\%)}
        & \multicolumn{1}{c}{(\%)}
        & \multicolumn{1}{c@{}}{(\%)} \\
        \midrule
        Shopping       & 53.4 & 65.7 & +12.3 & +26.4 \\
        Shopping Admin & 78.9 & 88.9 & +10.0 & +47.4 \\
        Reddit         & 56.8 & 63.6 &  +6.8 & +15.7 \\
        GitLab         & 67.9 & 76.1 &  +8.2 & +25.5 \\
        \midrule
        Mean & 64.3 & 73.6
             & \textbf{+9.3} & \textbf{+28.8} \\
        \bottomrule
    \end{tabular}%
    }
\end{wraptable}


\paragraph{Learning from more tries.}
This insight motivates Hippo to collect $N$ concurrent trajectories per task using the same policy and retrieved memory.
The different paths allow comparisons between successes and failures when both occur.
Hippo records surprise-triggered reflections in $L_1$.
A fix for one failure should not automatically become a general rule.
We should check whether the problem recurs, specify when the rule applies, and test it on other tasks.


\subsection{Number of rollouts}
\label{sec:rollout-count-analysis}
In our WebArena experiments, tasks receiving memories
from majority-success and majority-failure outcomes show average
absolute gains in success rate of $23.0\%$ and $13.0\%$,
respectively, compared with $3.3\%$ for the all-failure group.
To collect contrasting trajectories within a practical sampling
budget, we use eight rollouts per learning task, providing twice as many rollouts as $N=4$ at half
the sampling budget of $16$
(Appendix~\ref{sec:appendix_rollout_count}).












\HippoFinishWrap
\FloatBarrier
\subsection{Case Studies}

\begin{wrapfigure}{R}{0.42\textwidth}
    \centering
    \captionsetup{
        font=footnotesize,
        skip=3pt,
        justification=raggedright,
        singlelinecheck=false
    }

    \includegraphics[width=\linewidth]
        {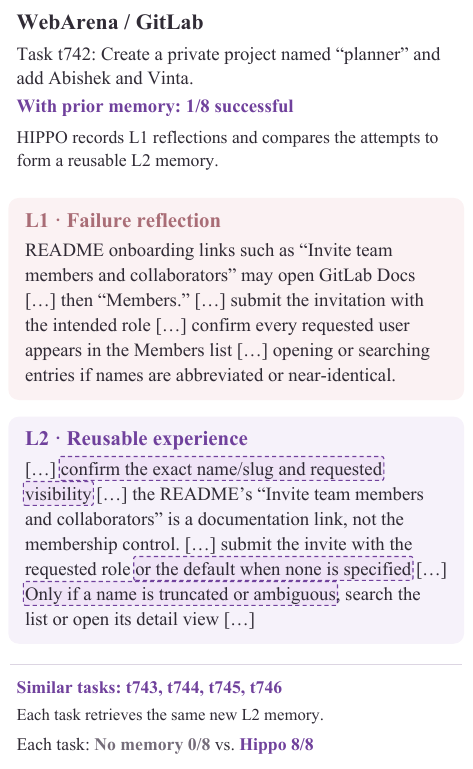}

    \captionof{figure}{
        Memory formation and reuse exmaple on WebArena.
Excerpts use [\ldots] for omissions; dashed boxes mark L2 additions and refinements.
    }
    \label{fig:case-gitlab}
\end{wrapfigure}

\citet{zhao2026faithful} find that replacing experience
with meaningless symbols (e.g., \texttt{\%\$\#\&}) sometimes does not
reduce task success, and identified overly general 
guidance as explanation.
Motivated by this, we examine what concrete guidance HIPPO
extracts from past attempts and how that guidance is reflected in subsequent execution.

Figure~\ref{fig:case-gitlab} highlights the value of consolidating
evidence across attempts. Repeated failures reveal a recurring error, while successful execution provides evidence for a workable alternative. HIPPO brings observations in a reusable
procedure with explicit applicability and completion criteria.

The L1/L2 hierarchy thus connects local diagnoses to guidance that
remains useful as task-specific details change. The resulting
experience specifies both how to act and what evidence is needed
to establish completion, with additional inspection conditioned
on observable ambiguity. These results illustrate how experience
formed from a largely unsuccessful batch can support reliable execution
on related tasks. Additional very detailed case studies and failure analyses are provided in
Appendix~\ref{app:learned-experiences}.

\subsection{Ablation Study}
\label{sec:ablation}

We evaluate ablations with GPT-5.6 Sol under the same
settings as the main experiments.


\noindent\textbf{Experience hierarchy.}
Without $L_2$, the agent retrieves $L_1$ reflections directly
and achieves $66.50\%$ on WebArena and $83.80\%$ on
SWE-bench Verified.
Without subsequent experience construction and checks
against existing memory, these local reflections may retain
task-specific details or provide inconsistent guidance
on later tasks.
Without $L_1$, we retain the judge and the $L_2$ engine
but omit explicit reflections. The engine builds $L_2$ experiences directly
from raw trajectories and judge feedback.

\HippoFinishWrap
\Needspace{145pt}
\begin{wraptable}{r}{0.38\textwidth}
\centering
\captionsetup{
    type=table,
    font=footnotesize,
    skip=4pt,
    justification=raggedright,
    singlelinecheck=false
}
\captionof{table}{Ablation study (\textit{w/o}) on WebArena and
SWE-bench Verified. \textit{w/o} means without.
$L_1$ denotes episodic reflections and $L_2$ denotes
consolidated experiences.}
\label{tab:ablation_hierarchy}

\fontsize{8.5}{10}\selectfont
\setlength{\tabcolsep}{2pt}
\renewcommand{\arraystretch}{1.03}
\begin{tabular*}{\linewidth}{@{\extracolsep{\fill}}lrr@{}}
\toprule
\textbf{Method}
& \shortstack{\textbf{WebArena}\\\textbf{(\%)}}
& \shortstack{\textbf{SWE-bench}\\\textbf{Verified (\%)}} \\
\midrule
No Memory      & 64.20 & 81.50 \\
w/o $L_2$      & 66.50 & 83.80 \\
w/o $L_1$      & 68.70 & 85.63 \\
\textbf{Hippo} & \textbf{72.90} & \textbf{88.50} \\
\bottomrule
\end{tabular*}

\end{wraptable}

\noindent This variant achieves $68.70\%$ and $85.63\%$, respectively.
Explicit reflections may help the engine focus on relevant
errors and their context before cross-rollout comparison.
Both variants perform worse than full Hippo at $72.90\%$
and $88.50\%$ in Table~\ref{tab:ablation_hierarchy}.
These results support reflection and subsequent
 construction of experiences.


\HippoFinishWrap
\Needspace{140pt}
\begin{wraptable}{r}{0.40\textwidth}
\centering
\captionsetup{
    type=table,
    font=footnotesize,
    skip=4pt,
    justification=raggedright,
    singlelinecheck=false
}
\captionof{table}{Ablation study on WebArena:
experience construction and memory updates.}
\label{tab:ablation_maintenance}

\fontsize{8.5}{10}\selectfont
\setlength{\tabcolsep}{2pt}
\renewcommand{\arraystretch}{1.03}
\begin{tabular*}{\linewidth}{@{\extracolsep{\fill}}lr@{}}
\toprule
\textbf{Method} & \shortstack{\textbf{Success}\\\textbf{(\%)}} \\
\midrule
Separate summaries + append only
    & 61.30 \\[1pt]
Compare rollouts + append only
    & 70.10 \\[1pt]
Compare rollouts + repeated rewrite
    & 54.57 \\[1pt]
\textbf{Hippo (add and delete)}
    & \textbf{72.90} \\
\bottomrule
\end{tabular*}

\end{wraptable}

\noindent\textbf{Comparing multiple attempts.}
We next examine how the $L_2$ engine constructs experience
(Table~\ref{tab:ablation_maintenance}).
Both variants use append-only memory updates.
Separate rollout summaries achieve $61.30\%$ success
on WebArena, whereas cross-rollout comparison achieves 70.10\%.
A lesson drawn from one attempt may reflect details
specific to that execution.
Comparison across attempts provides additional evidence
about which lessons to retain and when to apply them.
The strategies may also differ in the number and length
of stored entries, so this result evaluates the experience
construction strategy as a whole.


\noindent\textbf{Memory update strategies.}
With rollout comparison fixed, Hippo achieves 72.90\%,
compared with 70.10\% for append-only storage and 54.57\%
for repeated rewriting.
Append-only storage leaves earlier guidance unchanged,
even when new experience reveals a conflict.
Hippo checks when each experience applies and replaces
outdated guidance through explicit additions and deletions.
The repeated-rewrite variant also permits direct edits and
merges of existing entries.
Its lower score shows that broader freedom to revise memory
does not necessarily improve subsequent performance.
Appendix~\ref{app:memory_update_policies} details these policies.

\paragraph{Contribution of the surprise judge.}
Replacing the judge with just success/failure labels reduces
success rates by 1.8\%--8.5\% across the five WebArena
websites. We observe that answers such as \texttt{N/A} or \texttt{None}
can be marked correct despite an incomplete search or an
execution failure. These attempts can then enter memory as
successful experience and cause the agent to repeat the same
mistakes on later tasks.
Map contains the most such cases in our runs and shows the
largest decrease, at 8.5\%.
Detailed results and case studies are provided in
Appendix~\ref{app:judge_analysis}.

\par

\FloatBarrier
\HippoFinishWrap
\subsection{Efficiency Analysis.}
On WebArena, Hippo uses approximately 240--273 memory tokens per injection and 2,800 tokens per rollout
receiving memory.
With GPT-5.6 Sol, Hippo increases the success rate from
64.2\% to 72.9\% (+8.7 \%) over No Memory.
The resource records show that mean calls to the actor model
rise from 69.27 to 74.75 (+7.9\%), while cost increases
from \$1.93 to \$2.17 per task (+12.4\%).
Detailed comparisons and accounting scope are provided in
Appendix~\ref{app:efficiency}.

\section{Conclusion}

Hippo learns from multiple attempts and maintains reusable
experience without updating model parameters.
Experiments across three task settings show improved performance.
In our long-horizon evaluation, Hippo sustains gains with fewer
memory entries than the evaluated memory baselines,
whose performance eventually falls below No Memory.
Our retry analysis also motivates direct retries as a baseline
for future evaluations of self-evolving agents.
Future work includes reflection within longer tasks and
more efficient memory updates.
Limitations and future directions are discussed in
Appendix~\ref{sec:limitations}.




\FloatBarrier

\bibliography{iclr2027_conference}
\bibliographystyle{iclr2027_conference}

\clearpage
\appendix
\section{Appendix}

\subsection{Interaction steps.}
\label{sec:appendix_efficiency}
We analyze Hippo's rollout lengths to examine how much of the
30-step interaction budget is used in practice
(Table~\ref{tab:hippo_rollout_steps}).
On Shopping, Shopping Admin, GitLab, and Reddit, completed rollouts
use 8.12--9.40 steps on average, with 90th-percentile lengths of
15--20 steps.
Map has longer rollouts, with a mean of 13.39 steps and a
90th percentile at the 30-step limit.
Across websites, 2.8--10.3\% of completed rollouts reach this limit;
most therefore terminate before exhausting their step budget.

\begin{table}[!htbp]
\centering
\captionsetup{font=small,skip=5pt,width=0.80\linewidth}
\caption{Interaction steps of Hippo on WebArena.}
\label{tab:hippo_rollout_steps}
\begingroup
\fontsize{9}{11}\selectfont
\setlength{\tabcolsep}{4pt}
\renewcommand{\arraystretch}{1.1}
\begin{tabular}{@{}lrrrrr@{}}
\toprule
Website & \shortstack{Completed\\rollouts} & \shortstack{Mean\\steps} & \shortstack{Median\\steps} & \shortstack{P90\\steps} & \shortstack{Reached\\30 steps} \\
\midrule
Shopping
& 1,492 & 8.12 & 5 & 20 & 100 (6.7\%) \\
Shopping Admin
& 1,430 & 8.85 & 7 & 17 & 94 (6.6\%) \\
GitLab
& 1,430 & 9.40 & 8 & 19 & 40 (2.8\%) \\
Reddit
& 791 & 8.76 & 7 & 15 & 28 (3.5\%) \\
Map
& 836 & 13.39 & 12 & 30 & 86 (10.3\%) \\
\bottomrule
\end{tabular}
\par\vspace{4pt}
\begin{minipage}{0.80\linewidth}
\raggedright
\fontsize{9}{11}\selectfont
\textit{Note.} The interaction limit is 30 steps per rollout. Statistics cover completed rollouts regardless of success, including those that reach the limit; crashes are excluded. P90 denotes the 90th percentile. Percentages use the completed-rollout count for each website.
\end{minipage}
\endgroup
\end{table}

\section{Efficiency Analysis}
\label{app:efficiency}

We examine memory input, model calls, and an API cost proxy on
WebArena with GPT-5.6 Sol and eight rollouts per task. Per-task
statistics use 763 records for No Memory and 761 for Hippo, with
two Hippo task-level error cases excluded from the task count.
Memory input statistics cover 6,064 recorded Hippo rollouts,
of which 3,750 receive memory.

\paragraph{Memory input.}
Mean memory input ranges from 240 to 273 token equivalents per
injected step across websites (Table~\ref{tab:efficiency_tokens}).
Repeated use within a rollout increases cumulative input to an
overall mean of 2.8k token equivalents per injected rollout;
site-level p90 values range from 3.80k to 8.42k. We use four
characters per token equivalent and count each injection separately. The
resulting totals measure memory text supplied during execution,
rather than stored memory size or total API token usage.

\begin{table}[!htbp]
\centering
\captionsetup{font=small,skip=5pt,width=0.84\linewidth}
\caption{Memory input in Hippo.}
\label{tab:efficiency_tokens}
\begingroup
\fontsize{9}{11}\selectfont
\setlength{\tabcolsep}{3.5pt}
\renewcommand{\arraystretch}{1.1}
\begin{tabular}{@{}lrrrrrrr@{}}
\toprule
& \multicolumn{2}{c}{Rollouts} & \shortstack{Tokens\\per step} & \multicolumn{3}{c}{Per rollout (k tokens)} & Total \\
\cmidrule(lr){2-3}\cmidrule(lr){5-7}
Website & Injected & Recorded & Mean & Mean & Median & p90 & (M) \\
\midrule
Shopping       & 1,072 & 1,496 & 262 & 2.39 & 1.49 & 6.86 & 2.56 \\
Shopping Admin & 696   & 1,455 & 273 & 2.98 & 2.18 & 7.77 & 2.07 \\
GitLab         & 720   & 1,438 & 249 & 2.52 & 2.12 & 5.00 & 1.81 \\
Reddit         & 530   & 839   & 240 & 2.30 & 2.02 & 3.80 & 1.22 \\
Map            & 732   & 836   & 268 & 3.86 & 3.28 & 8.42 & 2.83 \\
\midrule
\textbf{Overall} & \textbf{3,750} & \textbf{6,064} & \textbf{260} & \textbf{2.80} & --- & --- & \textbf{10.5} \\
\bottomrule
\end{tabular}
\par\vspace{4pt}
\begin{minipage}{0.84\linewidth}
\raggedright
\fontsize{9}{11}\selectfont
\textit{Note.} Token equivalents equal character counts divided by four. Step and rollout statistics are conditional on memory injection. Cumulative values include repeated injections; pooled median and p90 are unavailable.
\end{minipage}
\endgroup
\end{table}

\paragraph{Execution and memory operations.}
Memory operations account for 4.1 of Hippo's 78.8 model calls per
task; the remaining 74.7 calls execute the agent. Relative to the
69.3 calls per task used by No Memory, the combined call count
increases by 13.8\%. Table~\ref{tab:efficiency_operations} separates
L1 reflection, L2 writing, and gating, and reports L1 invocations
alongside the reflections produced. These counts distinguish the
work spent processing memory from the number of entries written.

\begin{table}[!htbp]
\centering
\captionsetup{font=small,skip=5pt,width=0.80\linewidth}
\caption{Memory operations in Hippo.}
\label{tab:efficiency_operations}
\begingroup
\fontsize{9}{11}\selectfont
\setlength{\tabcolsep}{5pt}
\renewcommand{\arraystretch}{1.1}
\begin{tabular}{@{}lrrrrr@{}}
\toprule
& \multicolumn{2}{c}{L1 reflection} & L2 writing & Gating & Total \\
\cmidrule(lr){2-3}
Website & Calls & Written & Calls & Calls & Calls \\
\midrule
Shopping       & 630   & 560   & 103 & 186 & 919 \\
Shopping Admin & 500   & 454   & 84  & 179 & 763 \\
GitLab         & 360   & 314   & 65  & 178 & 603 \\
Reddit         & 153   & 142   & 43  & 104 & 300 \\
Map            & 332   & 303   & 74  & 108 & 514 \\
\midrule
\textbf{Overall} & \textbf{1,975} & \textbf{1,773} & \textbf{369} & \textbf{755} & \textbf{3,099} \\
\bottomrule
\end{tabular}
\par\vspace{4pt}
\begin{minipage}{0.80\linewidth}
\raggedright
\fontsize{9}{11}\selectfont
\textit{Note.} L1 invocations and produced reflections are separate quantities. Total calls sum L1, L2, and gating.
\end{minipage}
\endgroup
\end{table}

\paragraph{API cost proxy.}
We also report a monetary cost proxy because calls can differ in
their input and output lengths. It uses  USD
figures for both methods and covers task execution, trajectory
assessment, and memory operations, including injected memory text
(Table~\ref{tab:efficiency_cost}). The proxy averages \$2.17 per task
for Hippo and \$1.93 for No Memory. But, the relative increase is 12.4\%.

\begin{table}[!htbp]
\centering
\captionsetup{font=small,skip=5pt,width=0.80\linewidth}
\caption{API cost proxy in USD.}
\label{tab:efficiency_cost}
\begingroup
\fontsize{9}{11}\selectfont
\setlength{\tabcolsep}{4pt}
\renewcommand{\arraystretch}{1.1}
\begin{tabular}{@{}lrrrrrr@{}}
\toprule
& \multicolumn{3}{c}{No Memory} & \multicolumn{3}{c}{Hippo} \\
\cmidrule(lr){2-4}\cmidrule(lr){5-7}
Website & Tasks & Total (\$) & \$/task & Tasks & Total (\$) & \$/task \\
\midrule
Shopping       & 187 & 251.15 & 1.34 & 187 & 335.50 & 1.79 \\
Shopping Admin & 182 & 656.82 & 3.61 & 182 & 669.98 & 3.68 \\
GitLab         & 180 & 302.38 & 1.68 & 180 & 341.73 & 1.90 \\
Reddit         & 106 & 150.42 & 1.42 & 105 & 166.50 & 1.59 \\
Map            & 108 & 111.24 & 1.03 & 107 & 135.78 & 1.27 \\
\midrule
\textbf{Overall} & \textbf{763} & \textbf{1,472.02} & \textbf{1.93} & \textbf{761} & \textbf{1,649.50} & \textbf{2.17} \\
\bottomrule
\end{tabular}
\par\vspace{4pt}
\begin{minipage}{0.80\linewidth}
\raggedright
\fontsize{9}{11}\selectfont
\textit{Note.} The proxy includes all related model calls. Overall means pool the cost proxy and task counts across websites.
\end{minipage}
\endgroup
\end{table}

\subsection{Memory Retrieval and Selection}
\label{app:hippo_retrieval}

We evaluate memory selection on 752 recorded WebArena queries~\citep{zhou2024webarena}, using the same-site $L_2$ experiences available at each query's retrieval time.
The baselines are replayed on these historical banks and compared with Hippo's recorded selections, without rerunning the downstream agent.
An experience counts as a match if its source task shares the query's intent template.
Selection rate is measured on the 363 queries with a match available; correct abstention is measured on the remaining 389.

We compare Hippo with BM25~\citep{robertson2009bm25}, dense top-1 retrieval with BGE~\citep{xiao2024cpack}, hybrid sparse--dense ranking via reciprocal rank fusion~\citep{cormack2009reciprocal}, and similarity-threshold policies.
As shown in Table~\ref{tab:hippo_retrieval}, Hippo achieves the highest selection rate among these policies: 75.2\%, compared with 66.4\% for BM25 with experience content.
The paired gain is 8.8 percentage points, with a 95\% task-bootstrap confidence interval~\citep{efron1979bootstrap} of [4.1, 13.5].
When no match is available, Hippo abstains in 62.0\% of queries.
A cosine threshold of 0.65 raises correct abstention to 72.2\%, but reduces selection to 33.3\%. We also evaluate direct LLM selection from the full
historical memory bank. Its selection rate is 47.8\%,
compared with 75.2\% for Hippo.
This result supports a short candidate list before
the LLM selects an experience.

\begin{table}[!htbp]
\centering
\captionsetup{font=small,skip=5pt,width=0.86\linewidth}
\caption{Memory selection and abstention.}
\label{tab:hippo_retrieval}
\begingroup
\fontsize{9}{11}\selectfont
\setlength{\tabcolsep}{6pt}
\renewcommand{\arraystretch}{1.1}
\begin{tabular}{@{}lrr@{}}
\toprule
Policy & Selection $\uparrow$ & Correct abstention $\uparrow$ \\
\midrule
BM25                     & 62.8 &   1.3 \\
BM25 (+content)          & 66.4 &   0.8 \\
Dense top-1              & 59.8 &   0.0 \\
Hybrid (RRF)             & 65.3 &   0.0 \\
Dense + $\tau=0.65$      & 33.3 &  72.2 \\
Dense + $\tau=0.75$      &  7.4 &  99.5 \\
LLM-only (full memory bank) & 47.8 & n/a \\
\midrule
Hippo: cosine top-5 + LLM & \textbf{75.2} & 62.0 \\
\bottomrule
\end{tabular}
\par\vspace{4pt}
\begin{minipage}{0.86\linewidth}
\raggedright
\fontsize{9}{11}\selectfont
\textit{Note.} Relevance uses the same-template criterion. All values are percentages. BM25 (+content) also indexes the experience body; all dense policies use bge-small.
\end{minipage}
\endgroup
\end{table}

\begin{figure}[htbp]
\centering
\includegraphics[width=0.72\linewidth]{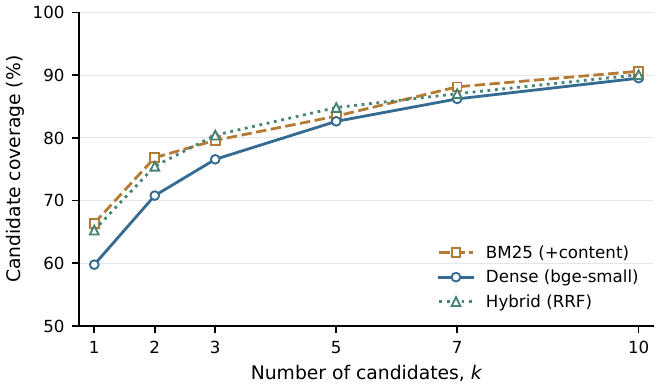}
\caption{Candidate coverage on the 363 queries with a same-template experience available.
Coverage is the fraction of queries whose top-$k$ candidates contain a match.
The curves measure candidate retrieval.
Among shortlist-based methods, only cosine top-5 was evaluated with the LLM selector.
Direct selection from the full memory bank was evaluated separately.}
\label{fig:hippo_retrieval_coverage}
\end{figure}

Retrieving several candidates gives the selector access to matches below rank one.
Among the 363 queries with a match available, dense top-1 selects one in 217 cases, while the top-5 contains one in 300 (Figure~\ref{fig:hippo_retrieval_coverage}).
The LLM selects a match in 273 of those 300 cases, a conditional selection rate of 91.0\%.
The resulting selection rate is 273/363, or 75.2\%; the remaining 90 queries comprise 63 without a match in the top-5 and 27 with a match that the LLM does not select.

Template agreement is a relevance proxy, and Hippo uses an additional LLM call.
The comparison therefore does not establish downstream gains at equal cost.
Other candidate generators and shortlist sizes were not tested with the same LLM, so their relative effectiveness remains open.
Within this scope, the results support cosine retrieval followed by LLM selection for identifying same-template experiences.

\section{Experimental Details}
\label{app:experimental_details}

\paragraph{Model use.}
Each Hippo run uses one model for task execution, trajectory assessment, and memory updates.
Model parameters remain fixed throughout the run.

\paragraph{SWE bench Verified.}
We use all 500 tasks in the \texttt{test} split of \texttt{princeton-nlp/SWE-bench\_Verified}.
Each attempt runs in a separate task container.
We use Epoch AI arm64 images where available.
Supplementary images or separate runs cover the remaining tasks.
For the main results, we evaluate all eight patches per task with the official \texttt{swebench} 4.1.0 harness.
We average the eight binary outcomes within each task, then across all 500 tasks. Each Hippo run starts with an empty memory bank.
The memory engine updates the bank only after all eight
attempts for a task have finished, so experiences from
that task become available only to subsequent tasks. All methods process the tasks in task-ID order,
with the same order across the three runs.

\paragraph{WebArena protocol.}
The main results use eight rollouts per task.
All methods and ablated variants use a maximum of
30 interaction steps per rollout, except Vanilla-IB,
which uses 50.
Tasks follow task-ID order, and all Hippo rollouts for a task use the same memory snapshot.
Hippo completes the rollouts and updates memory before the next task begins.
BrowserGym~\citep{drouin2024workarena,chezelles2025browsergym} calls the WebArena evaluator at each environment step, and the last returned reward defines each rollout's outcome.
We average success over the eight rollouts within each task, then across tasks. 

\paragraph{Mind2Web.}
We evaluate Mind2Web offline under teacher forcing, using
gold action histories and candidate sets constructed from
gold element labels.
For each attempt, we average step success within the task.
For the main results, we then average the eight attempt scores for each task, followed by an average across tasks. At each step, the page and execution context are reset,
while persistent memory is retained.
Memory also carries over across the three benchmark splits.
AWM, ReasoningBank, and Hippo use the same split order
and memory reset policy.

\paragraph{Long-horizon task construction.}
\label{app:long_horizon_tasks}
The task collection for the long-horizon experiment includes
coding tasks selected from SWE-bench Pro and processed
for evaluation.
All methods use the same task collection and follow
the same task order.

\paragraph{WebArena infrastructure.}
We host five WebArena websites on eight AWS EC2 \texttt{m6i.xlarge} instances.
Each instance has 4 vCPUs, 16 GB of memory, and a 1000 GiB \texttt{gp3} root volume with 3000 IOPS.
The Shopping, Shopping Admin, Reddit, and GitLab images occupy about 424 GB in total.
Concurrent rollouts use separate deployments.
Before each comparison condition, we restore the website containers from the baseline images.

\paragraph{Infrastructure costs.}
At the quoted deployment rates, the eight EC2 instances cost approximately USD~1.54 per hour.
Their eight storage volumes cost approximately USD~640 per month.
With continuous operation and a 30-day month, this corresponds to about USD~58.3 per day for compute and storage.
These estimates cover WebArena compute and storage and exclude model API calls and other charges.

\subsection{Long-horizon comparison with append-only memory}
\label{app:long_append_only}

We compare full Hippo with its append-only variant at the
400-task checkpoint.
The revisited set contains the first 50 tasks, while the
unseen set contains 50 fixed tasks.
Evaluation does not update the memory used for subsequent
learning.
As shown in Table~\ref{tab:long_append_only}, full Hippo
achieves higher final success on both sets.
The comparison supports the value of memory updates beyond
the addition of new experience.

\subsection{Success rates after learning tasks}
\begin{table}[htbp]
\centering
\begin{minipage}{0.72\linewidth}
\centering
\captionsetup{font=small,skip=5pt}
\caption{Success rates (\%) after 400 learning tasks,
evaluated on the first 50 tasks and a fixed set of
50 unseen tasks.}
\label{tab:long_append_only}
\small
\begin{tabular}{@{}lrr@{}}
\toprule
Method & Revisited & Unseen \\
\midrule
Append-only Hippo & 72.0 & 66.0 \\
Full Hippo & \textbf{86.0} & \textbf{82.0} \\
\bottomrule
\end{tabular}
\end{minipage}
\end{table}

\providecommand{\HippoCasePageBreak}{}
\providecommand{\HippoOmit}{\textup{[\ldots]}}
\providecommand{\HippoRole}[1]{\par\smallskip\noindent\textbf{#1.}\ }
\newtcolorbox{hippocase}[2][]{%
  title={#2}, sharp corners, boxrule=0.35pt,
  colframe=black!40, colback=black!1,
  colbacktitle=black!4, coltitle=black,
  fonttitle=\small\bfseries, fontupper=\small,
  boxsep=0pt, left=3mm, right=3mm, top=2mm, bottom=2mm,
  toptitle=1.8mm, bottomtitle=1.8mm, titlerule=0.25pt,
  before skip=4pt, after skip=4pt, #1}
\newtcolorbox{hippoprompt}[2][]{%
  breakable, lines before break=5,
  title after break={#2\space(continued)},
  title={#2}, sharp corners, boxrule=0.3pt,
  colframe=black!35, colback=black!1,
  colbacktitle=black!3, coltitle=black,
  fonttitle=\small\bfseries, fontupper=\small,
  boxsep=0pt, left=3mm, right=3mm, top=2mm, bottom=2mm,
  toptitle=1.5mm, bottomtitle=1.5mm, titlerule=0.2pt,
  before skip=7pt, after skip=7pt, #1}

\section{Judge Reliability and Contribution to Learning}
\label{app:judge_analysis}
\label{sec:appendix_judge_audit}
\label{app:surprise_feedback}

\subsection{Why outcome labels are not enough}
\label{app:surprise_motivation}

\textbf{Our WebArena runs reveal that an accepted response can come from
an incomplete attempt.} Among 412 accepted traces flagged by the judge as
fortuitous successes, 193 (46.8\%) have N/A-containing references, and 192
of these have no final response (Table~\ref{tab:judge_flagged_composition}).
These counts describe judge-flagged traces, not all WebArena tasks. A failed search or an early stop can thus
receive the same score as a careful check that finds no matching result.
If the accepted trace becomes a successful example for memory construction,
the agent may learn to repeat an unsupported decision.

Hippo uses the judge to assess the evidence behind accepted responses and
to diagnose failed attempts. The official score remains fixed: an incorrect
attempt is labeled \texttt{failure}, while an accepted attempt is labeled
\texttt{genuine} or \texttt{fluke}. When the evidence for an accepted attempt
is ambiguous, the judge defaults to \texttt{genuine}. These labels guide
reflection and experience construction without changing benchmark scores.

\begin{table}[!htbp]
\centering
\captionsetup{font=small,skip=5pt,width=0.82\linewidth}
\caption{Composition of the recorded WebArena pool.}
\label{tab:judge_flagged_composition}
\begingroup
\fontsize{9}{11}\selectfont
\setlength{\tabcolsep}{5pt}
\renewcommand{\arraystretch}{1.1}
\begin{tabular}{@{}lrrrrr@{}}
\toprule
Website & Accepted & Flagged & \shortstack{\texttt{N/A}\\reference} & \shortstack{No\\answer} & Both \\
\midrule
Shopping & 1,698 & 99 & 83 & 99 & 83 \\
Shopping Admin & 1,859 & 44 & 0 & 32 & 0 \\
GitLab & 2,103 & 103 & 31 & 100 & 30 \\
Reddit & 1,278 & 26 & 23 & 26 & 23 \\
Map & 999 & 140 & 56 & 88 & 56 \\
\midrule
Total & 7,937 & 412 & 193 & 345 & 192 \\
\bottomrule
\end{tabular}
\par\vspace{4pt}
\begin{minipage}{0.82\linewidth}
\raggedright
\fontsize{9}{11}\selectfont
\textit{Note.} Accepted counts all traces accepted by the evaluator in the full recorded pool. Flagged counts those subsequently labeled fortuitous successes by the judge. The last three columns are counts within the flagged set.
\end{minipage}
\endgroup
\end{table}

\subsection{Feedback available during memory construction}
\label{app:surprise_protocol}

\paragraph{WebArena}
Within Hippo, the judge receives the reference answer and official outcome
along with the task, final response, and trace. The $L_1$ and $L_2$ writers
receive its labels and diagnoses, but no separate reference-answer field
(Table~\ref{tab:hippo_feedback_access}). A diagnosis can still contain
reference-derived values or assessment criteria. The actor receives only
selected past experience in addition to its normal task inputs; it has no
direct access to the current task's reference, reward, or judge reason.

\begin{table}[htbp]
\centering
\captionsetup{font=small,skip=5pt,width=0.90\linewidth}
\caption{Feedback and experience available to each Hippo component.}
\label{tab:hippo_feedback_access}
\small
\setlength{\tabcolsep}{5pt}
\renewcommand{\arraystretch}{1.15}
\begin{tabular}{@{}l>{\raggedright\arraybackslash}p{0.66\linewidth}@{}}
\toprule
Component & Feedback and experience \\
\midrule
Judge & Official outcome, reference, and observed trace \\
$L_1$ writer & Judge label and diagnosis; individual trace \\
$L_2$ writer & Outcomes and diagnoses; $L_1$ or genuine-success traces \\
Actor & Selected past experience; no direct current-task feedback \\
\bottomrule
\end{tabular}
\end{table}

In our WebArena comparison, AWM and ReasoningBank also receive official
outcomes and reference answers during memory construction. We enable
MaTTS for ReasoningBank. The three methods therefore have access to
reference information, although they use it differently. This is an
adaptation of their original online settings, which use model-based
outcome judgments~\citep{wang2025awm,ouyang2026reasoningbank}.

The original evaluators also serve a different purpose. AWM adopts
the evaluator of \citet{pan2024autonomous}, while ReasoningBank uses its agent backbone to
predict success or failure. ReasoningBank reports 72.7\% accuracy for
Gemini-2.5-flash on WebArena-Shopping~\citep{ouyang2026reasoningbank}.
Hippo instead takes the official result as given and assesses the trace
or diagnoses the failure.

\paragraph{Mind2Web.}
We evaluate Mind2Web as an offline stream under teacher forcing,
using gold action histories and candidate sets constructed from
gold element labels.
At each step, the page and execution context are reset,
while Hippo's persistent $L_2$ memory is retained.
The memory bank also carries over across the Cross-Task,
Cross-Website, and Cross-Domain splits.
AWM, ReasoningBank, and Hippo use the same split order
and memory reset policy.
The reported scores therefore describe adaptation along a shared
memory stream.
The judge is not told which candidate is correct.
For each attempt, we average step success within the task.
We then average the eight attempt scores for each task,
followed by an average across tasks.

\paragraph{SWE-bench Verified.}
The judge is label-free. It receives the issue, execution trace,
and candidate patch, without the reference patch or official
outcome label. It returns two binary fields, \texttt{success}
and \texttt{verified}, for memory construction.
The reported resolution rate is computed separately with
the official SWE-bench harness.

\subsection{Effect of the judge}
\label{app:judge_effect}

We replace the judge with official success/failure labels. This removes
both the genuine/fluke distinction and the textual diagnoses, including
those for failed attempts. Success rates fall on all five WebArena
websites, by 1.8--8.5 percentage points
(Table~\ref{tab:judge_module}). The comparison measures the combined
contribution of trajectory assessment and diagnostic feedback.

Map contains the most N/A or None cases in our runs and has the largest
decrease, at 8.5 percentage points. Such cases make the distinction between
a verified absence and an incomplete attempt especially relevant.
The case in Section~\ref{app:surprise-cases} illustrates this distinction;
Section~\ref{app:case-failed-source} shows how failure diagnoses can support
experience construction.

\begin{table}[!htbp]
\centering
\captionsetup{font=small,skip=5pt,width=0.64\linewidth}
\caption{Effect of removing the judge.}
\label{tab:judge_ablation}
\label{tab:judge_module}
\begingroup
\fontsize{9}{11}\selectfont
\setlength{\tabcolsep}{7pt}
\renewcommand{\arraystretch}{1.1}
\begin{tabular}{@{}lr@{}}
\toprule
Website & Decrease (pp) \\
\midrule
Shopping       & 3.5 \\
Shopping Admin & 1.8 \\
GitLab         & 3.5 \\
Reddit         & 1.8 \\
Map            & 8.5 \\
\bottomrule
\end{tabular}
\par\vspace{4pt}
\begin{minipage}{0.64\linewidth}
\raggedright
\fontsize{9}{11}\selectfont
\textit{Note.} Success-rate decrease relative to full Hippo, in percentage points (pp).
\end{minipage}
\endgroup
\end{table}

\subsection{Agreement with human annotations}
Two human annotators assess 200 evaluator-accepted traces, with 50 traces from each of four actor models.
For each of the five websites, the audit includes 20 randomly sampled traces labeled as genuine successes and 20 labeled as fortuitous successes by the judge.
The annotators agree on 195 of the 200 traces (97.5\%; Cohen's $\kappa = 0.95$).
After excluding five disagreements and one inconclusive case, the judge agrees with the human labels on 183 of the remaining 194 traces (94.3\%), with Cohen's $\kappa = 0.89$.
Among the 94 retained traces flagged as fortuitous successes, humans confirm 86 (91.5\%).
Table~\ref{tab:judge_reliability} reports the results by website.
Agreement varies across sites: the judge matches all retained human labels on Shopping and Reddit, while agreement on Map is 84.6\% ($\kappa = 0.69$).
All metrics describe the stratified audit sample, with judge--human comparisons restricted to conclusive human agreement.

\begin{table}[!htbp]
\centering
\captionsetup{font=small,skip=5pt,width=0.86\linewidth}
\caption{Agreement in the human audit of judge labels.}
\label{tab:judge_reliability}
\begingroup
\fontsize{9}{11}\selectfont
\setlength{\tabcolsep}{3.5pt}
\renewcommand{\arraystretch}{1.1}
\begin{tabular}{@{}lrrrrrr@{}}
\toprule
& \multicolumn{2}{c}{Human--human} & \multicolumn{3}{c}{Judge--human} & \shortstack{Flag\\precision} \\
\cmidrule(lr){2-3}\cmidrule(lr){4-6}
Website & Agree (\%) & $\kappa$ & $n$ & Agree (\%) & $\kappa$ & (\%) \\
\midrule
Shopping & 100.0 & 1.00 & 40 & 100.0 & 1.00 & 100.0 \\
Shopping Admin & 90.0 & 0.79 & 36 & 91.7 & 0.83 & 81.3 \\
GitLab & 100.0 & 1.00 & 39 & 94.9 & 0.90 & 94.7 \\
Reddit & 100.0 & 1.00 & 40 & 100.0 & 1.00 & 100.0 \\
Map & 97.5 & 0.95 & 39 & 84.6 & 0.69 & 78.9 \\
\midrule
\textbf{Pooled} & \textbf{97.5} & \textbf{0.95} & \textbf{194} & \textbf{94.3} & \textbf{0.89} & \textbf{91.5} \\
\bottomrule
\end{tabular}
\par\vspace{4pt}
\begin{minipage}{0.86\linewidth}
\raggedright
\fontsize{9}{11}\selectfont
\textit{Note.} Each website contributes 40 traces, split equally between the judge's two labels. Human--human metrics use all 200 traces; judge--human metrics use the $n=194$ traces with matching, conclusive human labels. Flag precision is the percentage of retained judge-flagged traces confirmed by humans. Metrics are unweighted descriptions of the stratified sample; pooled $\kappa$ is computed from pooled labels.
\end{minipage}
\endgroup
\end{table}

\paragraph{Agreement measures.}
We report raw agreement and Cohen's $\kappa$~\citep{cohen1960agreement} separately for the two human annotators and for the judge against conclusive human consensus.
For the binary judge--human comparison, let $n=\mathrm{TP}+\mathrm{FP}+\mathrm{FN}+\mathrm{TN}$.
The observed and chance-expected agreements are
\begin{equation}
p_o = \frac{\mathrm{TP}+\mathrm{TN}}{n}, \qquad
p_e = \frac{(\mathrm{TP}+\mathrm{FP})(\mathrm{TP}+\mathrm{FN})
+ (\mathrm{FN}+\mathrm{TN})(\mathrm{FP}+\mathrm{TN})}{n^2},
\end{equation}
and $\kappa=(p_o-p_e)/(1-p_e)$.
Using the pooled counts in Table~\ref{tab:judge_audit_counts}, $p_o=183/194$, $p_e=18{,}866/37{,}636$, and $\kappa=0.8863$, reported as 0.89.
The pooled coefficient is computed from pooled counts rather than an average of site-level coefficients.
Flag precision is $\mathrm{TP}/(\mathrm{TP}+\mathrm{FP})$ within the retained audit sample.
These statistics do not correct for the stratified sampling design or the exclusion of unresolved labels.

\begin{table}[!htbp]
\centering
\captionsetup{font=small,skip=5pt,width=0.84\linewidth}
\caption{Human audit counts for evaluator-accepted WebArena traces.}
\label{tab:judge_audit_counts}
\begingroup
\fontsize{9}{11}\selectfont
\setlength{\tabcolsep}{5pt}
\renewcommand{\arraystretch}{1.1}
\begin{tabular}{@{}lrrrrrr@{}}
\toprule
Website & Audited & Retained & TP & FP & FN & TN \\
\midrule
Shopping & 40 & 40 & 20 & 0 & 0 & 20 \\
Shopping Admin & 40 & 36 & 13 & 3 & 0 & 20 \\
GitLab & 40 & 39 & 18 & 1 & 1 & 19 \\
Reddit & 40 & 40 & 20 & 0 & 0 & 20 \\
Map & 40 & 39 & 15 & 4 & 2 & 18 \\
\midrule
Total & 200 & 194 & 86 & 8 & 3 & 97 \\
\bottomrule
\end{tabular}
\par\vspace{4pt}
\begin{minipage}{0.84\linewidth}
\raggedright
\fontsize{9}{11}\selectfont
\textit{Note.} Each website contributes 20 judge-genuine and 20 judge-flagged traces. Retained traces have matching, conclusive human labels. Fortuitous success is the positive class: TP and FP count retained judge-flagged traces, while FN and TN count retained judge-genuine traces. Counts describe the stratified audit sample.
\end{minipage}
\endgroup
\end{table}

\subsection{Repeatability on the same input}
Repeated judging produces consistent labels on 24 of 25 checked traces (96.0\%).
This check measures consistency across repeated evaluations of the same input, separately from agreement with human labels.

\subsection{Cases of incomplete verification}
\label{app:surprise-cases}

\subsubsection{An accepted response without completed verification}

In the first case, the reference is N/A. The agent takes one action and
returns no final answer, yet the fuzzy-match evaluator accepts the response.
The trace indicates that a required check is still unfinished, so the judge
labels the attempt as \texttt{fluke}. The record has no task identifier.

\noindent
\begin{minipage}[t]{0.486\linewidth}
\begin{hippocase}[equal height group=HippoAcceptedCase]{A1\quad What the evaluator accepted}
The agent inspected a project list but did not complete its check of the
user's contribution activity.

\medskip
\begin{tabularx}{\linewidth}{@{}l>{\raggedright\arraybackslash}X@{}}
\toprule
Field & Recorded value \\
\midrule
Reference & \texttt{N/A} (fuzzy match) \\
Final response & \texttt{""} \\
Actions & 1 \\
Official result & Correct \\
Judge label & \texttt{fluke} \\
\bottomrule
\end{tabularx}
\end{hippocase}
\end{minipage}\hfill
\begin{minipage}[t]{0.486\linewidth}
\begin{hippocase}[equal height group=HippoAcceptedCase]{A2\quad Why the judge flagged it}
\textbf{Judge reason (verbatim)}

``The agent produced no final answer. Although it inspected the project list
and noticed no displayed repository exceeded 100 stars, it explicitly
concluded that the profile contribution activity still needed inspection and
never completed that verification. The empty response only happened to align
with the N/A reference.''
\end{hippocase}
\end{minipage}

The \texttt{fluke} label marks the attempt for reflection. The record
contains no experience generated from this trace.

\subsubsection{How an unsupported procedure can enter memory}
An earlier GitLab experiment illustrates how an accepted outcome can enter memory without a verified procedure behind it.
Task t168 asks for repositories to which the user contributed that have more than 100 stars; its reference answer is \texttt{N/A}.
In the earlier implementation, the experience writer could access the reference answer and produced a lesson that advised returning \texttt{N/A} if no repository exceeded the threshold.
Three related tasks (t169--t171) subsequently retrieved this lesson and returned \texttt{N/A}, although their reference answers named repositories.
Each task failed in all eight attempts.

A later implementation both excluded fortuitous successes from the contrast used for induction and withheld the raw reference-answer field from the writer.
It produced an experience that instructed the agent to open Projects, select Yours, sort by Most stars, and verify the star counts.
The same three related tasks each succeeded in all eight attempts; an additional related task, t172, also succeeded in all eight attempts.
This historical comparison illustrates how different inputs to experience induction can accompany different stored lessons and subsequent behavior.
Because both trace filtering and direct access to the
reference-answer field changed, it does not isolate the causal contribution of the surprise judge or evaluate the current implementation.

\section{Rollout Outcomes and Budget}
\label{sec:appendix_rollout_count}

\paragraph{Setup and measures.}
We analyze 761 WebArena tasks with gpt-5.6-sol at $N=8$.
Of these tasks, 471 received experience and 290 did not.
We group the 471 tasks by the outcomes of the rollouts used to produce their experience.
The balanced group has equal counts of successes and failures.
Both majority groups include at least one success and one failure.
Let $n_g$ be the task count of group $g$ and $\Delta_g$ its mean score gain.
All groups use the same reference condition.
The group's contribution $C_g$ to the overall score is
\begin{equation}
    \underbrace{C_g}_{\text{overall contribution}}
    =
    \underbrace{\frac{n_g}{M}}_{\text{task coverage}}
    \times\underbrace{\Delta_g}_{\text{within-group gain}},
    \qquad M=761,
    \label{eq:rollout-gain-normalization}
\end{equation}
where $C_g$ and $\Delta_g$ are in percentage points (pp).

\begin{table}[!htbp]
\centering
\captionsetup{font=small,skip=5pt,width=0.74\linewidth}
\caption{Gains by experience source at $N=8$.}
\label{tab:rollout-group-effects}
\begingroup
\fontsize{9}{11}\selectfont
\setlength{\tabcolsep}{7pt}
\renewcommand{\arraystretch}{1.1}
\begin{tabular}{@{}lrrr@{}}
\toprule
Experience source & Tasks & $\Delta_g$ (pp) & $C_g$ (pp) \\
\midrule
All failure             & 183 & $+3.3$ & $+0.80$ \\
Majority failure        & 140 & $+13.0$ & $+2.39$ \\
Balanced                & 24  & $+55.8$ & $+1.76$ \\
Majority success        & 124 & $+23.0$ & $+3.75$ \\
\bottomrule
\end{tabular}
\par\vspace{4pt}
\begin{minipage}{0.74\linewidth}
\raggedright
\fontsize{9}{11}\selectfont
\textit{Note.} Both gain columns are in percentage points. Group means are rounded to one decimal place.
\end{minipage}
\endgroup
\end{table}

\paragraph{Gain and coverage.}
The all-failure group has the smallest mean gain (Table~\ref{tab:rollout-group-effects}).
The balanced group has the largest, but covers only 24 tasks and adds $1.76$ pp to the overall score.
The two majority groups have smaller gains per task, but cover 264 tasks and add $6.14$ pp in total.

\paragraph{Rollout budget.}
With $N=1$, we cannot compare success and failure within a task.
With $N=2$, such a comparison requires one success and one failure.
The larger choices, $N=4,8,16$, also allow unequal counts.
We choose $N=8$ because it allows more attempts than $N=4$ and needs half as many as $N=16$.
These results do not show which $N$ gives the best performance or the largest gain per token.

\subsection{Metrics for Repeated Attempts}
\label{sec:retry_metrics}

We compare two success measures for the No Memory agent
with GPT-5.6 Sol on four WebArena websites:
Shopping, Shopping Admin, Reddit, and GitLab.
For website $d$, let $p_d$ denote the mean per-attempt
success rate reported in the main table.
Let $a_{8,d}$ denote the fraction of tasks solved at least
once within eight attempts, without memory updates.

With both rates expressed as probabilities, we compute
the absolute gain $\Delta_d$ in percentage points and
the normalized gain $G_d$ as
\begin{equation}
\label{eq:retry_gains}
    \Delta_d = 100(a_{8,d}-p_d),
    \qquad
    G_d = 100\frac{a_{8,d}-p_d}{1-p_d}.
\end{equation}
For $p_d<1$, $G_d$ expresses the gain as a percentage
of the mean per-attempt failure rate.

We compute each metric separately for each website
and then average the four website-level values equally.
The resulting mean per-attempt and at-least-once success
rates are $64.3\%$ and $73.6\%$, respectively.
The mean absolute gain is $9.3$ percentage points.
The website-level normalized gains are
$26.4\%$, $47.4\%$, $15.7\%$, and $25.5\%$,
giving a mean of $28.8\%$.

The normalized gain compares the two success measures.
It does not track which tasks failed on the first attempt
and succeeded on a later attempt.

\section{Qualitative Analysis of Learned Experiences}
\label{app:learned-experiences}

We examine five cases that illustrate how Hippo turns recorded reflections and trajectory evidence into experiences that can be retrieved on later tasks. The cases cover conflicting recommendations, transfer between task templates, learning when all source attempts fail, the consolidation of an execution procedure, and reuse across capacity requirements. For Cases 1--4, each box presents selected L1 passages together with the complete title, applicability description, and procedural content of the resulting L2. The surrounding text explains the source evidence, the aspect of the method illustrated, and the observed behavior on later tasks. All counts are successful attempts out of eight within the corresponding run. Source results precede formation of the displayed L2 and should not be attributed to that new experience.

\noindent\textit{Presentation of memory records.} Memory wording is retained apart from sentence punctuation and menu-path notation. Semicolons are replaced with sentence breaks, and menu paths use slashes. Bracketed ellipses indicate omissions from L1. For Cases 1--4, L2 applicability and procedure fields are shown in full.

\subsection{Case 1. Reconciling reflections through explicit applicability}
\label{app:case-status}

Task t62 asks which customer has completed the most orders over the full history. HIPPO retrieves no prior experience on this task and succeeds in 4/8 attempts, compared with 6/8 without memory. The trajectories expose a distinction between obtaining the right name and using a data source that respects the question. Some successful attempts aggregate the general Customer Order Count report, while another uses orders restricted to Complete. The recorded L1 reflections also differ. Some recommend aggregating the report across years, whereas another explicitly requires status-filtered orders. HIPPO subsequently forms the L2 shown below.

\Needspace{10\baselineskip}
\begin{hippoexperience}{Experience 1. WebArena / Shopping Admin (formed after t62)}
\textbf{L1 reflection (excerpt): Aggregate the full Order Count report by customer}

Open REPORTS, then Customers / Order Count. Set a realistic full-history range, choose Year in ``Show By'' to avoid oversized daily reports, and select Refresh. Read every populated interval and sum the Orders value for every repeated customer name across all rows and years, then compare the resulting customer totals and report every tie. [\ldots{}]

\textbf{L1 reflection (excerpt): Count completed orders from Sales Orders}

Open Sales / Orders, expose the full historical grid, filter Status to Complete/Completed, and inspect every page or export all matching rows before grouping by the exact full Customer Name shown. [\ldots{}]

\smallskip\hrule\smallskip

\textbf{L2 experience: Use the status-filtered grid and aggregate every occurrence}

\textit{Applicability.} Applies to historical customer rankings, especially when the question restricts orders by status. The Reports / Customers / Order Count report is suitable only when all statuses qualify or its scope is independently verified.

\textit{Procedure.} For status-specific counts, use Sales / Orders, open Filters, apply the required Status, and ensure every matching record is visible across all pages or an export. Do not trust Order Count as a look-alike because it has no status filter. Group every exact full customer name consistently across the complete result---including repeated rows and intervals---compare final totals, include every tie, distinguish near-identical names, and open View if a grid value is abbreviated before answering only from visibly confirmed data.
\end{hippoexperience}

\paragraph{What this case illustrates.} HIPPO consolidates differing L1 recommendations into guidance with explicit applicability. The L2 preserves aggregation and tie handling while requiring status-filtered data for status-specific questions. This illustrates how consolidation retains task constraints across competing recommendations. The source judge also accepts correct answers from the general report, so the case supports conditional guidance rather than reliable detection of spurious success.

\paragraph{Reuse and observed effect.} On t63, which asks for the second-most completed orders, HIPPO retrieves the new L2 and achieves 8/8 successes versus 2/8 without memory. A representative attempt checks all 153 Complete orders and returns Adam Garcia, Michael Nguyen, and Sarah Miller, tied at eight orders each. Both tasks share template 276.

\subsection{Case 2. Transferring a verification procedure across task templates}
\label{app:case-range}

Task t126 asks for the price range of Canon photo printers and belongs to template 159. With an older price-range memory, HIPPO succeeds in 7/8 attempts, compared with 8/8 without memory. The single failed HIPPO attempt already finds the correct endpoints but does not submit a final answer. Successful source attempts also demonstrate how to exclude irrelevant products. The resulting experience preserves the working search and verification procedure together with the reminder to stop and answer once the endpoints are established. Thus, the batch contains useful information even though most source attempts already succeed.

\Needspace{10\baselineskip}
\begin{hippoexperience}{Experience 2. WebArena / Shopping (formed after t126)}
\textbf{L1 reflection (excerpt): Verify the complete range, then answer immediately}

Open ``Advanced Search,'' constrain ``Product Name'' with the most reliable brand or identifying term, increase ``Show'' to the largest page size, sort by ``Product Name,'' and inspect every result page. Treat cables, parts, supplies, cases, and near-identical nonqualifying variants as look-alike traps. [\ldots{}] once both visible endpoints are verified across the complete set, stop navigating and explicitly return the price range.

\smallskip\hrule\smallskip

\textbf{L2 experience: Verify the complete product range, then answer}

\textit{Applicability.} Use when a shopping task asks for a price range or other extrema across products. Do not use a broad catalog scan when a single product detail page already answers the question.

\textit{Procedure.} Open ``Advanced Search,'' constrain ``Product Name'' with the brand or identifying product term, use the largest ``Show'' option, and inspect every results page or sort by price. Exclude accessories, parts, supplies, and products that merely mention the brand. Read the full names and prices directly, opening product details if the grid abbreviates them, account for every qualifying or tied row, and once both visible endpoints are verified, stop navigating and explicitly provide the requested range---do not omit the final answer.
\end{hippoexperience}

\paragraph{What this case illustrates.} The L2 retains a product-verification procedure and the single failed attempt's reminder to submit an answer once both endpoints are confirmed. Its scope covers product ranges across brands. Reuse from template 159 to template 370 illustrates procedural transfer within one shopping site, even though the source failure concerns answer submission and the target failures concern product qualification.

\paragraph{Reuse and observed effect.} Tasks t226 and t228 both retrieve this L2. HIPPO achieves 8/8 on each, compared with 0/8 and 1/8 without memory, respectively. On t226, a representative attempt excludes a Polk audio bundle that only includes an Amazon Basics cable and reports \$5.49--\$375.19. On t228, the runs return the Sephora range of \$18.18--\$94.99.

\subsection{Case 3. Forming useful experience without a successful source attempt}
\label{app:case-failed-source}

Task t141 asks how much was spent on food-related shopping during March 2023. Both arms achieve 0/8, and no memory is retrieved for the HIPPO attempts. The recorded failure diagnoses identify omitted category-related items and missing attributable shipping. L1 reflections preserve these corrections, and HIPPO forms an L2 that combines complete order-history inspection, category coverage, and attributable charges. There is no successful source demonstration in this batch. The experience instead draws on the information supplied by failed attempts and their diagnoses.

\Needspace{10\baselineskip}
\begin{hippoexperience}{Experience 3. WebArena / Shopping (formed after t141)}
\textbf{L1 reflection (excerpt): Use full order details and include attributable costs}

From ``My Account,'' open ``My Orders'' rather than relying on the incomplete ``Recent Orders'' table, review every relevant page and date row, then use each row's ``View Order'' link to inspect ``Items Ordered'' and ``Shipping \& Handling.'' Enumerate every qualifying product---including clearly related supplies, decorations, or accessories rather than only consumables---read its full displayed name, variant, quantity, and subtotal, and include the shipping attributable to each qualifying item when the order details establish a per-item charge. [\ldots{}]

\smallskip\hrule\smallskip

\textbf{L2 experience: Audit complete orders and allocate related shipping}

\textit{Applicability.} Use for category-spend questions based on purchase history, especially when orders mix qualifying and unrelated merchandise. Do not rely on the limited ``Recent Orders'' panel or interpret a category as consumables only.

\textit{Procedure.} Open ``My Account'' / ``My Orders,'' review all pages for the requested period, and select ``View Order'' on every qualifying row. In ``Items Ordered,'' enumerate every visibly category-related product---including accessories, decorations, and supplies---using full names and displayed subtotals, then add each item's attributable share of ``Shipping \& Handling'' based on the visible per-item or per-unit pattern and reconcile the result with the order totals.
\end{hippoexperience}

\paragraph{What this case illustrates.} HIPPO forms reusable guidance from an entirely failed batch by combining category coverage and shipping checks. The diagnoses provide reference-informed corrections, so the learning signal exceeds binary reward. Transfer remains conditional: t142 retrieves the same experience but achieves 0/8, as does the baseline. Its equal shipping allocation is rejected, showing that the experience's allocation assumptions still require verification.

\paragraph{Reuse and observed effect.}
Task \texttt{t143} asks about home-decoration spending on January 29, 2023.
The selector chooses this experience from five candidates (index 0;
cosine similarity 0.585). All eight Hippo attempts receive the same
optional hint, recorded as 746 characters. The instruction tells the actor
to use it only when it clearly fits the current task.
All eight no-memory attempts return \$260.69, while all eight Hippo
attempts return \$265.69, giving 0/8 versus 8/8 success.
A separate diagnosis for no-memory rollout 0 on \texttt{t143} attributes
the \$5 difference to the item's share of a \$10 shipping charge.
That later diagnosis is not the source of the earlier \texttt{t141}
experience. Section~\ref{app:surprise-prompt-excerpts} provides the
recorded prompt excerpts for these roles.

\subsection{Case 4. Consolidating an execution procedure and its completion checks}
\label{app:case-inventory}

Task t501 asks the agent to mark all Taurus Elements Shell products and variants out of stock. HIPPO retrieves an older experience about verifying salable inventory and succeeds in 1/8 attempts, compared with 0/8 without memory. Failures occur at several stages. Some attempts cannot select all matching rows, others fail to perform the required stock update, and others stop when an asynchronous operation has only been submitted. Several attempts also mistake a pre-existing success notification for confirmation of the current task. The L1 reflections record corrections at these different stages, and the new L2 assembles them into the procedure below.

\Needspace{10\baselineskip}
\begin{hippoexperience}{Experience 4. WebArena / Shopping Admin (formed after t501)}
\textbf{L1 reflection (excerpt): Verified bulk stock update}

[\ldots{}] Select the complete filtered set using the header Options menu's Select All when the adjacent checkbox is obstructed, confirm the selected count, then use Actions to open the inventory/source-stock update, set stock status to out of stock, and save or confirm. [\ldots{}]

\textbf{L1 reflection (excerpt): Confirm queued bulk stock updates}

[\ldots{}] ``Message is added to queue'' confirms submission only: wait for a new job-specific success notification, ignoring older System Messages, then refresh and open matching product detail views as needed to verify every Stock Availability field visibly shows the requested value rather than inferring it from quantity, salable quantity, or product status.

\smallskip\hrule\smallskip

\textbf{L2 experience: Select all, update Stock Availability, and respect the queue}

\textit{Applicability.} Bulk-changing Stock Availability for every matching product and variant in Catalog / Products. Not changing product enablement status.

\textit{Procedure.} Filter the Products grid, inspect every matching parent and variant, use Options / Select All, confirm the selected count, then choose Actions / Update attributes / Advanced Inventory, check Change beside Stock Availability, choose the requested value, and Save. ``Change status'' and the grid's Status column control enablement, not inventory. If ``Message is added to queue'' appears, wait for a new operation-specific success notice before verification, ignoring pre-existing System Messages. Only then refresh and open product details as needed to read the actual Stock Status, rather than inferring it from quantity or salable quantity.
\end{hippoexperience}

\paragraph{What this case illustrates.} Consolidation organizes L1 corrections into a procedure that distinguishes inventory from product enablement and submission from completion. Execution remains imperfect: one failed t502 attempt selects all 16 products and submits the update, then reports completion while it is still queued. Some successful episodes terminate at saving without a later queue check. The case shows useful procedural guidance alongside incomplete compliance.

\paragraph{Reuse and observed effect.} Tasks t502 and t504 retrieve the new L2 for Gobi HeatTec Tee and Selene Yoga Hoodie products. HIPPO achieves 4/8 and 6/8, compared with 0/8 and 2/8 without memory, respectively. Successful attempts select the matching products and use Advanced Inventory to set Stock Availability to Out of Stock.

\subsection{Case 5. Reuse across capacity requirements}
\label{app:capacity_reuse}

Task \texttt{t158} asks for the best storage option for
11 Nintendo Switch game cards.
Hippo forms an $L_2$ experience titled
``Search broadly, then verify every qualifying product.''
On \texttt{t161}, which asks for storage for 23 cards,
Hippo retrieves the experience from \texttt{t158}.
On this target task, Hippo succeeds in 8/8 attempts,
compared with 0/8 for No Memory.
This case illustrates reuse of experience when the capacity
requirement changes within the same task template.

\paragraph{Relation to the overall method.} Together, these cases illustrate complementary functions of the experience hierarchy. L1 preserves the diagnoses associated with individual attempts, including different local recommendations, while L2 provides a retrievable procedure with an explicit scope. The outcomes show where those procedures are reused successfully and where their assumptions or execution remain incomplete. These are qualitative examples with paired task outcomes, rather than an isolated comparison of L2 retrieval against direct L1 retrieval. They explain the role of the design without assigning the entire observed improvement to any single clause or component.

\Needspace{10\baselineskip}
\section{Recorded Prompt Excerpts}
\label{app:surprise-prompt-excerpts}

The excerpts below document the inputs and outputs for the cases in
Sections~\ref{app:surprise-cases} and~\ref{app:case-failed-source}.
Bracketed ellipses mark editorial omissions; plain ellipses preserve
omissions in the supplied records. Compact traces and interaction
histories are not reproduced.
The separate no-memory diagnosis for \texttt{t143} illustrates the judge's
feedback format. It is not the source of the earlier \texttt{t141}
experience.

\begin{hippoprompt}{Judge input and output\quad \texttt{t143}, no-memory rollout 0}
\textbf{Recorded user fields}

\texttt{TASK:} How much I spent on home decoration shopping during 1/29/2023

\texttt{REFERENCE ANSWER (ground truth):}

\texttt{\{"must\_include": ["265.69"]\}}

\texttt{OFFICIAL GRADER VERDICT: INCORRECT}

\texttt{AGENT FINAL ANSWER: \$260.69}

\textbf{Trace (omitted).} Per-step thoughts, actions, and page text.

\textbf{Judge output (verbatim reason; outcome: \texttt{failure}).}
``The agent correctly identified the \$260.69 home-decoration item but
reported only its merchandise subtotal. It omitted the item's \$5 share of
the \$10 shipping charge across the two ordered items; adding that yields
the required \$265.69.''

\end{hippoprompt}

\subsection{Writers: reflection and experience construction}

The $L_1$ writer uses an individual trace and its diagnosis. The $L_2$
writer then compares the outcomes and the available reflections or
success traces. The following two inputs both come from \texttt{t141}.

\begin{hippoprompt}{L1 writer\quad \texttt{t141}, memory-arm rollout 0; system excerpts}
You are an expert web-navigation analyst distilling a lesson from ONE failed
rollout. It FAILED: The agent undercounted the mixed March 11 order. It
included only the \$11.43 cornbread mix and missed the \$7.99 cake-topper item
as food-related, while also failing to include the applicable share of that
order's \$25 shipping. \ldots

Extract lessons that would prevent this failure on future tasks on this
site. \HippoOmit

A single attempt cannot tell a bad route from a good route walked badly:
do NOT write blanket prohibitions (``never use X'', ``page Y is unreliable'')
from one trajectory --- state what to VERIFY on that route instead. \HippoOmit

NEVER copy task-specific values (product names, search terms, dates,
numbers). \HippoOmit

- Return at most 1 item --- the single most important, reusable lesson;
[] only if nothing reusable was learned.
\end{hippoprompt}

\begin{hippoprompt}{L2 writer\quad \texttt{t141}; system and user excerpts}
\textbf{System.} An agent made N independent PARALLEL rollouts of the SAME
web task. For each rollout you see its verdict (OK/WRONG, with the judge's
reason) and EITHER the lesson distilled from it (its L1) OR, for a genuine
success we want to learn from, its full TRACE.
Look across ALL N outcomes and the success/failure ratio, then write ONE
higher-level lesson: \HippoOmit

\textbf{Excerpt note.} The all-failure instruction is omitted here.
The prompt also includes ``NEVER copy
task-specific values (product names, search terms, dates, numbers).''

\textbf{User.} \texttt{SITE: shopping}

\texttt{TASK:} How much I spent on food-related shopping during March 2023

8 rollouts, 0 succeeded:

--- rollout 1 [WRONG] (The agent undercounted the mixed March 11 order.
\HippoOmit)

\quad L1 lesson: Audit mixed orders and allocate shipping: From My Account,
select ``View All'' \ldots

--- rollout 2 [WRONG] (The agent undercounted by treating only the cornbread
mix as food-related in the mixed order and using its \$11.43 merchandise
subtotal without shipping \ldots{} It also needed to include the \$7.99 cake
topper \ldots)

\quad L1 lesson: Use full order details and include attributable costs:
\ldots

--- rollout 3 [WRONG] (\ldots{} omitted the \$5-per-item shipping attributable
to each of the three relevant products \ldots)

\quad L1 lesson: Audit full order details and attributable costs: \ldots

\textbf{Excerpt note.} The other five rollout entries are omitted.
\end{hippoprompt}

\subsection{Selector and actor: experience use on a later task}

The selector receives the current task and candidate experiences. It can
choose one or return $-1$ to select none. The actor receives the selected
experience as an optional hint.

\begin{hippoprompt}{Applicability gate\quad \texttt{t143}; system and candidate excerpts}
\textbf{System.} Given the CURRENT web task and a numbered list of candidate
past lessons, pick the ONE most APPLICABLE lesson. It must genuinely fit
this task's goal and the controls/pages it needs --- matching topic words is
NOT enough \HippoOmit{} If NONE genuinely applies, return -1 --- injecting
nothing is better than a misleading lesson.

Respond JSON \texttt{\{"idx": int\}} (the 0-based index, or -1).

\textbf{Candidate 0.} [0] Audit complete orders and allocate related
shipping: Open ``My Account'' $\rightarrow$ ``My Orders,'' review all pages
\ldots

\textbf{Output.} The other four candidates are omitted.
The gate returns \texttt{\{"idx": 0\}}; cosine similarity is 0.585.
\end{hippoprompt}

\begin{hippoprompt}{Actor\quad \texttt{t143}; shared optional-hint excerpts}
Optional hints from past tasks on this site --- use one ONLY if it clearly
fits the current task, and prefer the cheapest sufficient evidence. Do not
add steps or chase a procedure a hint describes if you can already answer
directly and correctly:

[strategy] Audit complete orders and allocate related shipping: Open
``My Account'' $\rightarrow$ ``My Orders,'' review all pages for the requested
period, and select ``View Order'' on every qualifying row. \HippoOmit{}
then add each item's attributable share of ``Shipping \& Handling'' based
on the visible per-item or per-unit pattern and reconcile the result with
the order totals. \HippoOmit

\textbf{Use.} All eight rollouts receive this optional experience.
The interaction history is omitted.
\end{hippoprompt}
\subsection{Memory update policies}
\label{app:memory_update_policies}

We compare three policies for incorporating new experience
after cross-rollout comparison.

\paragraph{Append only.}
Each new experience is added to memory.
Existing entries remain unchanged, without removal or integration.

\paragraph{Repeated rewrite.}
The model receives existing memories and a new observation.
For each old entry, it can retain the text, revise it to add
detail or correct an error, merge it with semantically
duplicate entries, or delete it.
It can also add facts that existing entries do not cover.
An entry can remain unchanged; the policy does not require
every entry to be rewritten at every update.
The update instruction is:

\begin{quote}
\small
Given the existing memories and a new observation, decide
for each old entry whether to \textbf{keep} it,
\textbf{rewrite} it (fill in detail or correct errors),
\textbf{merge} it with semantically duplicate entries into
a single item, or \textbf{delete} it (falsified, stale, or
fully redundant), and add any facts from the new observation
that no existing entry covers.
\end{quote}

\paragraph{Hippo.}
Hippo compares new experience with related memories.
It discards duplicate candidates and resolves overlaps or
conflicts through explicit conditions of use.
When new guidance replaces an existing entry, Hippo adds
the replacement and deletes the old entry.
Entries outside the deletion set retain their original text.
This policy allows prior guidance to change while preserving
entries that the update does not replace.

\section{Limitations}
\label{sec:limitations}

Hippo incurs additional inference and interaction costs from
multiple rollouts and memory updates.
KV-cache reuse in SGLang~\citep{zheng2024sglang} and
vLLM~\citep{vllm2026prefix} can avoid recomputing identical
prompt prefixes across rollouts, reducing redundant inference computation.
Specialized models such as Jev~\citep{jev2026} could handle
judgment labels, while a generative model would still produce
textual diagnoses.
We have not evaluated Jev or quantified the end-to-end savings
from KV-cache reuse.

Hippo uses the same base model for task execution,
trajectory assessment, and memory updates.
Shared blind spots may allow incorrect or overly general lessons
to persist in memory.
A stronger model could construct and revise experience
for a smaller model to apply during execution.
We have not evaluated whether such experience transfers reliably
or whether its reuse offsets the added construction cost.

Our long-horizon evaluation studies memory updates across tasks;
it does not establish performance on exceptionally long individual tasks.
Longer trajectories may strain context budgets and make it harder
to retain early decisions and dependencies across stages.
Repeating complete rollouts would also increase the cost of collecting experience.
Reflection at intermediate stages and hierarchical consolidation
are possible extensions, but whether they preserve enough evidence
to derive reliable experience remains untested.

\end{document}